\documentclass[3p,twocolumn]{elsarticle}

\usepackage{hyperref}

\usepackage{amsmath}
\usepackage{amssymb}
\usepackage{graphicx}
\usepackage{tikz}
\usepackage{subfig}
\usepackage{rotating}
\usepackage{multirow}
\usepackage{url}
\usepackage{float}
\usepackage{color}

\definecolor{myred}{rgb}{.8,.0,.0}
\definecolor{mygreen}{rgb}{.0,.4,.0}
\definecolor{myblue}{rgb}{.0,.0,.8}

\journal{Medical Image Analysis}

\begin{document}

\begin{frontmatter}

\title{Transfer Learning by Asymmetric Image Weighting\\for Segmentation across Scanners}

\address[bigr]{Biomedical Imaging Group Rotterdam, Depts. Radiology and Medical Informatics, Erasmus Medical Center, Rotterdam, The Netherlands}
\address[prlab]{Pattern Recognition Laboratory, Dept. Intelligent Systems, Delft University of Technology, Delft, The Netherlands}
\address[brain]{Dept. Epidemiology and Radiology, Erasmus Medical Center, Rotterdam, The Netherlands}
\address[diku]{The Image Section, Dept. Computer Science, University of Copenhagen, Copenhagen, Denmark}

\author[bigr,prlab]{Veronika Cheplygina\corref{cor1}}
\ead{v.cheplygina@tue.nl}
\cortext[cor1]{Corresponding author. This research was performed while Veronika Cheplygina was with the Biomedical Imaging Group Rotterdam, Erasmus Medical Center, The Netherlands. She is now with the Medical Image Analysis group, Eindhoven University of Technology, The Netherlands. }

\author[bigr]{Annegreet van Opbroek}

\author[brain]{M. Arfan Ikram}

\author[brain]{Meike W. Vernooij}

\author[bigr,diku]{Marleen~de~Bruijne}

\begin{abstract}

Supervised learning has been very successful for automatic segmentation of images from a single scanner. However, several papers report deteriorated performances when using classifiers trained on images from one scanner to segment images from other scanners. We propose a transfer learning classifier that adapts to differences between training and test images. This method uses a weighted ensemble of classifiers trained on individual images. The weight of each classifier is determined by the similarity between its training image and the test image.

We examine three unsupervised similarity measures, which can be used in scenarios where no labeled data from a newly introduced scanner or scanning protocol is available. The measures are based on a divergence, a bag distance, and on estimating the labels with a clustering procedure. These measures are asymmetric. We study whether the asymmetry can improve classification. Out of the three similarity measures, the bag similarity measure is the most robust across different studies and achieves excellent results on four brain tissue segmentation datasets and three white matter lesion segmentation datasets, acquired at different centers and with different scanners and scanning protocols. We show that the asymmetry can indeed be informative, and that computing the similarity from the test image to the training images is more appropriate than the opposite direction.
\end{abstract}


\end{frontmatter}



\section{Introduction}\label{sec:intro}

Manual biomedical image segmentation is time-consuming and subject to intra- and interexpert variability, and thus in recent years a lot of advances have been made to automate this process. Because of its good performance, supervised voxelwise classification~\cite{opbroek2014transfer,van2015weighting,zikic2014encoding,zikic2014classifier,anbeek2005probabilistic,geremia2011spatial,steenwijk2013accurate,boer2009white,ithapu2014extracting}, where manually labeled images are used to train supervised classifiers, has been used successfully in many applications. These include brain tissue (BT) segmentation and white matter lesion (WML) segmentation~\cite{van2015weighting,anbeek2005probabilistic,geremia2011spatial,steenwijk2013accurate,boer2009white,ithapu2014extracting}.

However, supervised classifiers need labeled data that is representative of the target data that needs to be segmented in order to be successful. In multi-center studies or longitudinal studies, differences in scanners or scanning protocols can influence the appearance of voxels, causing the classifier to deteriorate when applied to data from a different center. For example, \cite{steenwijk2013accurate} show on two independent datasets that their WML classifier performs well in each dataset separately, but that performance degrades substantially when the classifier is trained on one dataset and tested on the other. In a study of WML segmentation with three datasets from different centers, \cite{van2015weighting} shows a large gap in performance between a classifier trained on same-center images, and classifiers trained on different-center images, despite using intensity normalization.


Most WML segmentation approaches in the literature do not address the multi-center problem. A recent survey~\cite{garcia2013review} of WML segmentation, shows that out of 47 surveyed papers, only 13 papers used multi-center data, and 11 of those only used the datasets from the MS lesion challenge~\cite{styner20083d}. The survey therefore states robustness in multi-center datasets as one of the remaining challenges for automatic WML segmentation. Even when multi-center data is used, evaluation may still assume the presence of labeled training data from each center. For example, \cite{geremia2011spatial} uses the two MS lesion challenge datasets, which have 10 scans each, in a joint 3-fold cross-validation. This means that at each fold, the classifier is trained on 14 subjects, which necessarily includes subjects from both centers.

In BT segmentation multi-scanner images are sometimes addressed with target-specific atlas selection in multi-atlas label propagation~\cite{zikic2014classifier,lombaert2014laplacian}. Although these papers do not specifically focus on images with different feature distributions, selecting atlases that are similar to the test image could help to alleviate the differences between the training and the test data. However, there are some details which make the methods less suitable for multi-center situations. Zikic et al~\cite{zikic2014classifier} use class probabilities based on a model of intensities of \emph{all} images as additional features. Differences in feature distributions of the images could produce an inaccurate model, and the features would therefore introduce additional class overlap.


\emph{Transfer learning}~\cite{pan2010survey} techniques can be employed in order to explicitly deal with the differences between source and target data. Such methods have only recently started to emerge in medical imaging applications. These approaches frequently rely on a small amount of labeled target data (\cite{opbroek2014transfer,becker2014domain,cheng2012domain,conjeti2016supervised,goetz2016dalsa}, to name a few), or can be unsupervised with respect to the target~\cite{van2015weighting,heimann2014real}, which is favorable for tasks where annotation is costly. In the latter case, typically the transfer is achieved by weighing the training samples such that the differences between training and target data are minimized. For example, \cite{van2015weighting} weight the training images such that a divergence, such as Kullback-Leibler (KL), between the training and test distributions is minimized. These image weights are then used to weight the samples before training a support vector machine (SVM).

We propose to approach voxelwise classification by a similarity-weighted ensemble of random forests~\cite{breiman2001random} (RF). The approach is general and can be applied to any segmentation task. The classifiers are trained \emph{only once}, each on a different source image. For a target image, the classifier outputs are fused by weighted averaging, where the weights are determined by the similarity of the source image and the target image. The method does not require any labeled data acquired with the test conditions, is computationally efficient and can be readily applied to novel target images. The method is conceptually similar to multi-atlas segmentation, but has an explicit focus on different training and test distributions, which is currently underexplored in the literature. Furthermore, in medical image segmentation, little attention has been paid to asymmetric similarity measures. Such measures have shown to be informative in classification tasks in pattern recognition applications~\cite{pkekalska2006non,cheplygina2016asymmetric}, but, to the best of our knowledge, have not been investigated in the context of similarity-weighted ensembles. \textbf{The novelty of our contribution lies in the comparison of different unsupervised asymmetric similarity measures, which allow for on-the-fly addition of training or testing data, and insights into how to best deal with asymmetric similarity measures in brain MR segmentation.}

This paper builds upon a preliminary conference paper~\cite{cheplygina2016asymmetric}, where we applied our method to BT segmentation. In the present work, we also apply the method to WML segmentation. In addition, we investigate how different parameters affect the classifier performance, and provide insight into why asymmetry should be considered. We outperform previous benchmark results on four (BT) and three (WML) datasets acquired under different conditions. On the WML task, our method is also able to outperform a same-study classifier trained on only a few images, acquired with the same conditions as the test data.



\section{Materials and Methods}

\subsection{Brain Tissue Segmentation Data}

We use the brain tissue segmentation dataset from \cite{van2015weighting}, which includes 56 manually segmented MR brain images from healthy young adults and elderly:

\begin{itemize}
    \item 6 T1-weighted images from the Rotterdam Scan Study (RSS)~\cite{ikram2011rotterdam} acquired with a 1.5T GE scanner at 0.49$\times$0.49$\times$0.8 mm$^3$ resolution. We refer to this set of images as RSS1.

    \item 12 half-Fourier acquisition single-shot turbo spin echo (HASTE) images scanned with a HASTE-Odd protocol from the Rotterdam Scan Study, acquired with a 1.5T Siemens scanner at 1.25$\times$1$\times$1 mm$^3$ resolution. These HASTE-Odd images resemble inverted T1 images, and were therefore inverted during the preprocessing of the data. We refer to this set of images as RSS2.

    \item 18 T1-weighted images from the Internet Brain Segmentation Repository (IBSR)~\cite{ibsr}, acquired with multiple unknown scanners, at resolutions ranging from 0.84$\times$0.84$\times$1.5 mm$^3$ to 1$\times$1$\times$1.5 mm$^3$. We refer to this set of images as IBSR1.

    \item 20 T1-weighted images from the IBSR~\cite{ibsr}, of which 10 are acquired with a 1.5T Siemens scanner and 10 are acquired with a 1.5T GE scanner, in all cases at 1$\times$1.3$\times$1 mm$^3$ resolution. We refer to this set of images as IBSR2.
\end{itemize}

The scans of RSS1 and RSS2 are of older subjects, while the scans of IBSR are of young adults. The age of the subjects influences the class priors of the tissues encountered in the images: RSS subjects have relatively more cerebrospinal fluid (CSF) and less gray matter (GM) than young adults.


\subsection{White Matter Lesion Data}
\begin{figure*}
    \centering
     \includegraphics[width=0.15\linewidth,height=3cm]{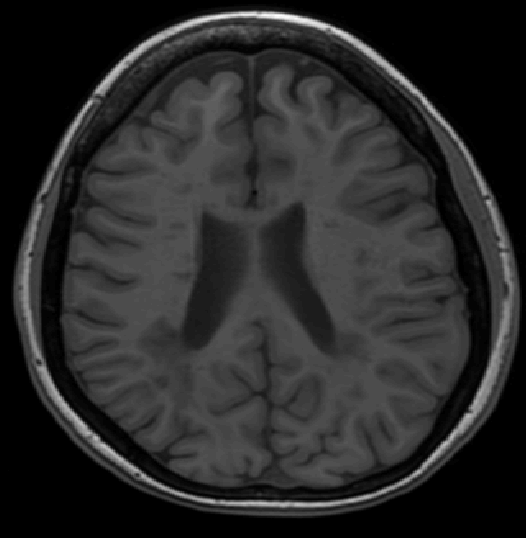}
     \includegraphics[width=0.15\linewidth,height=3cm]{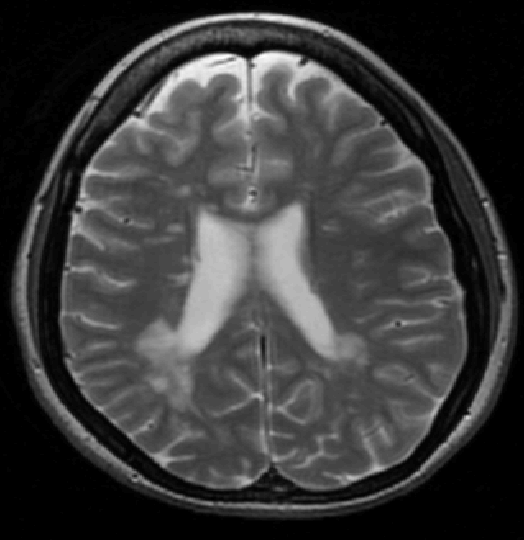}
     \includegraphics[width=0.15\linewidth,height=3cm]{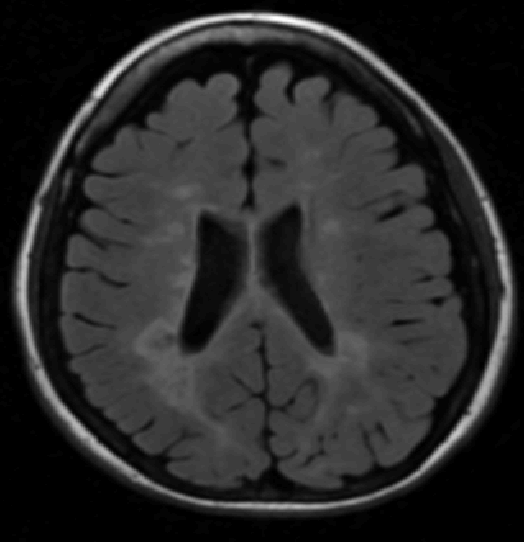}
     \includegraphics[width=0.15\linewidth,height=3cm]{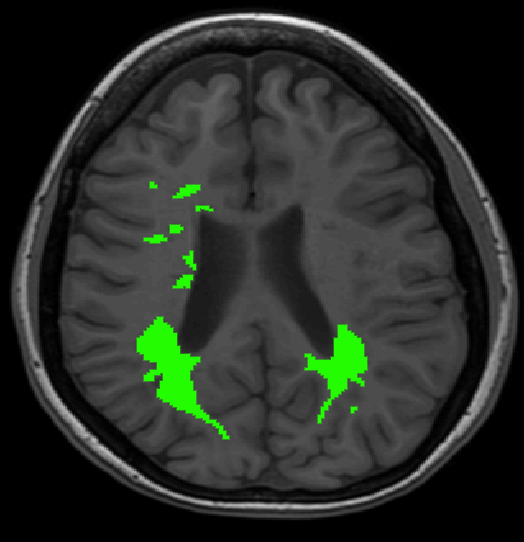}

     \includegraphics[width=0.15\linewidth,height=3cm]{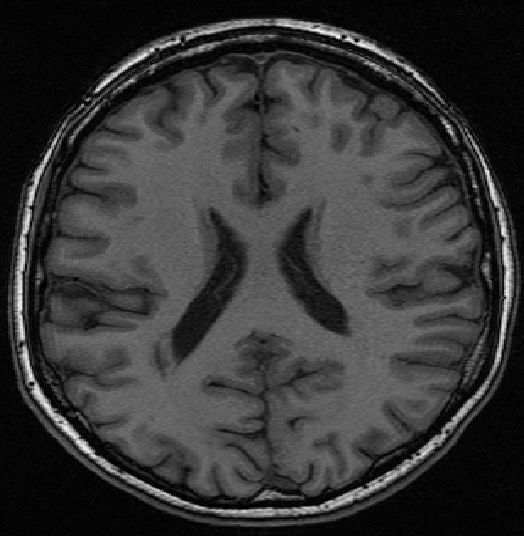}
     \includegraphics[width=0.15\linewidth,height=3cm]{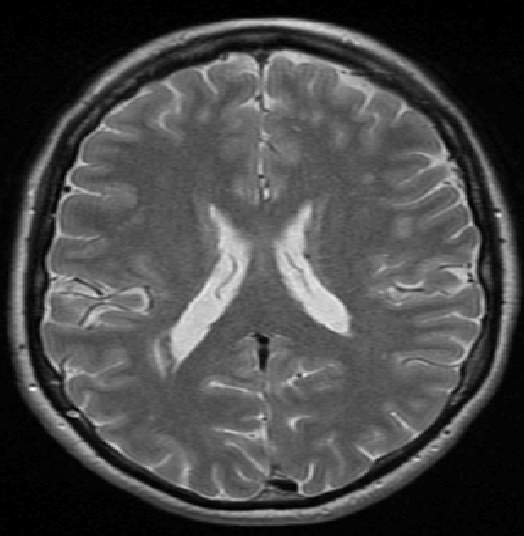}
     \includegraphics[width=0.15\linewidth,height=3cm]{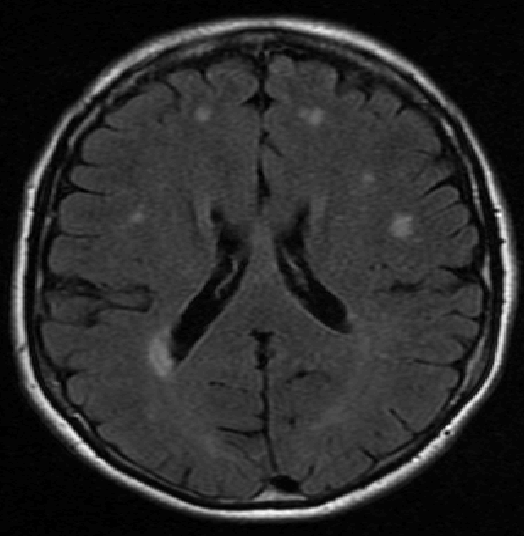}
     \includegraphics[width=0.15\linewidth,height=3cm]{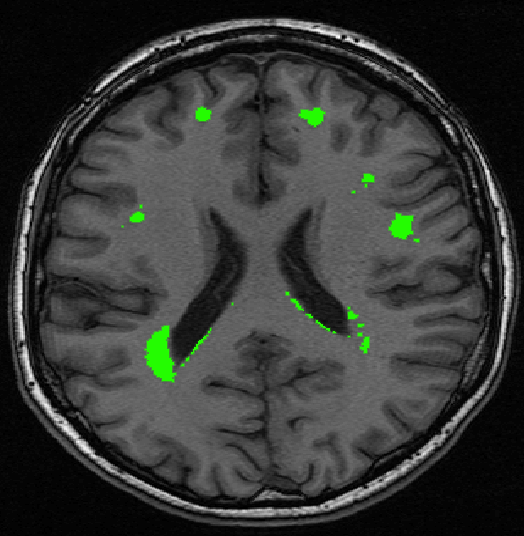}

     \includegraphics[width=0.15\linewidth,height=3cm]{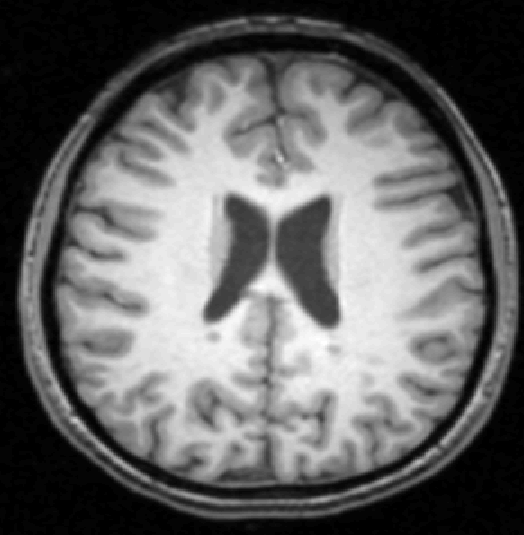}
     \includegraphics[width=0.15\linewidth,height=3cm]{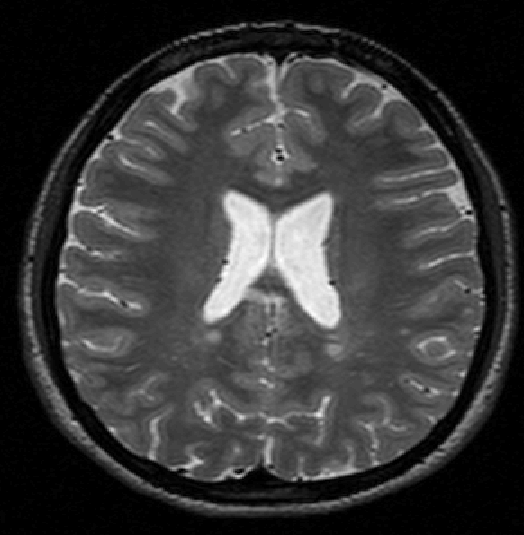}
     \includegraphics[width=0.15\linewidth,height=3cm]{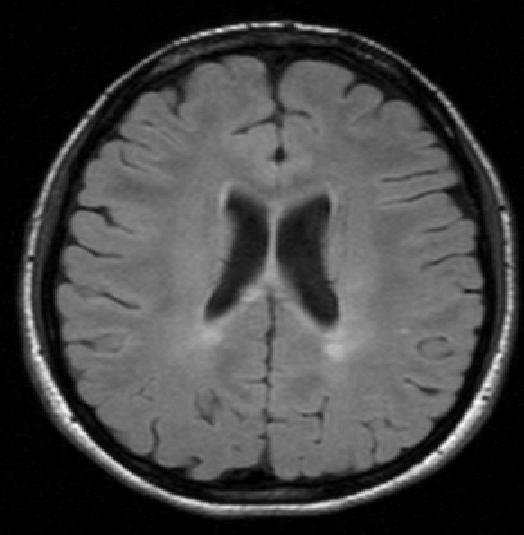}
     \includegraphics[width=0.15\linewidth,height=3cm]{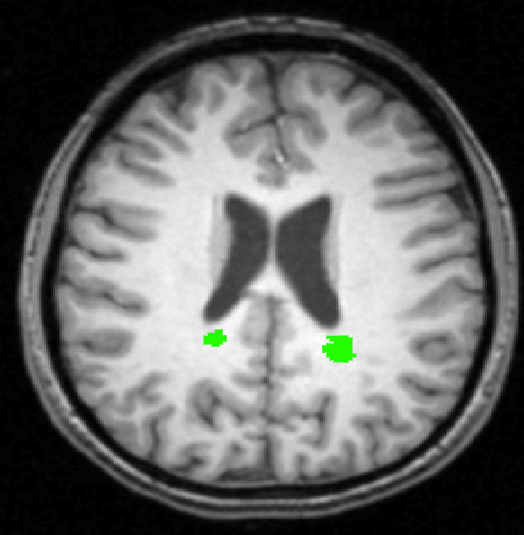}

    \caption{Examples of slices from the three different modalities (T1, T2 or PD, FLAIR) and manual annotations (overlaid in green on the T1 image) from three datasets (CHB, RSS and UNC).}
    \label{fig:examples}
\end{figure*}

We use images from three different studies (see Fig.~\ref{fig:examples} for examples of slices):
\begin{itemize}
\item 10 MS patients from the MS Lesion Challenge~\cite{styner20083d} scanned at the Children's Hospital of Boston (CHB), scanned with T1, T2 and FLAIR at 0.5$\times$0.5$\times$0.5mm resolution.

\item 10 MS patients from the MS Lesion Challenge~\cite{styner20083d} scanned at the University of North Carolina (UNC), scanned with T1, T2 and FLAIR at 0.5$\times$0.5$\times$0.5mm resolution.

\item 20 healthy elderly subjects with WML from the RSS~\cite{ikram2011rotterdam,ikram2015rotterdam}, scanned with T1, PD and FLAIR sequences at 0.49$\times$.0.49$\times$0.8mm resolution (T1 and PD) and  0.49x0.49x2.5 resolution (FLAIR). Because PD images of RSS appear similar to the T2 images of CHB and UNC, these modalities are treated to be the same.
\end{itemize}

Here again the differences between study populations influence the class priors. On average, the percentage of voxels that are lesions are 1.6\%, 2.6\% and 0.2\% in CHB, RSS and UNC respectively. The differences between subjects also vary: these are relatively small for CHB and UNC, but very large for RSS. In RSS, the subject with the least lesion voxels has only 0.08\%, while the patient with the most lesion voxels has 14.3\%.


\subsection{Image Normalization and Feature Extraction}

We approach segmentation by voxelwise classification. We therefore represent each voxel by a vector of features describing the appearance of the voxel. Prior to feature extraction, initial image normalization was performed. This normalization included bias-field correction with the N4 method~\cite{tustison2010n4itk} (both BT and WML data), inversion of HASTE-Odd images (BT only) and normalizing the voxel intensities by [4,96]-th percentile range matching to the interval [0,1] (both BT and WML data). For BT data, range matching was performed inside manually annotated brain masks. For WML, when scans of modalities were obtained at different resolutions, they were co-registered to the T1 scan.
For WML, range matching was performed inside manually annotated brain masks (RSS) or masks generated with with BET~\cite{smith2002fast} (CHB and UNC).

\begin{figure}%
\centering
\includegraphics[width=0.85\columnwidth]{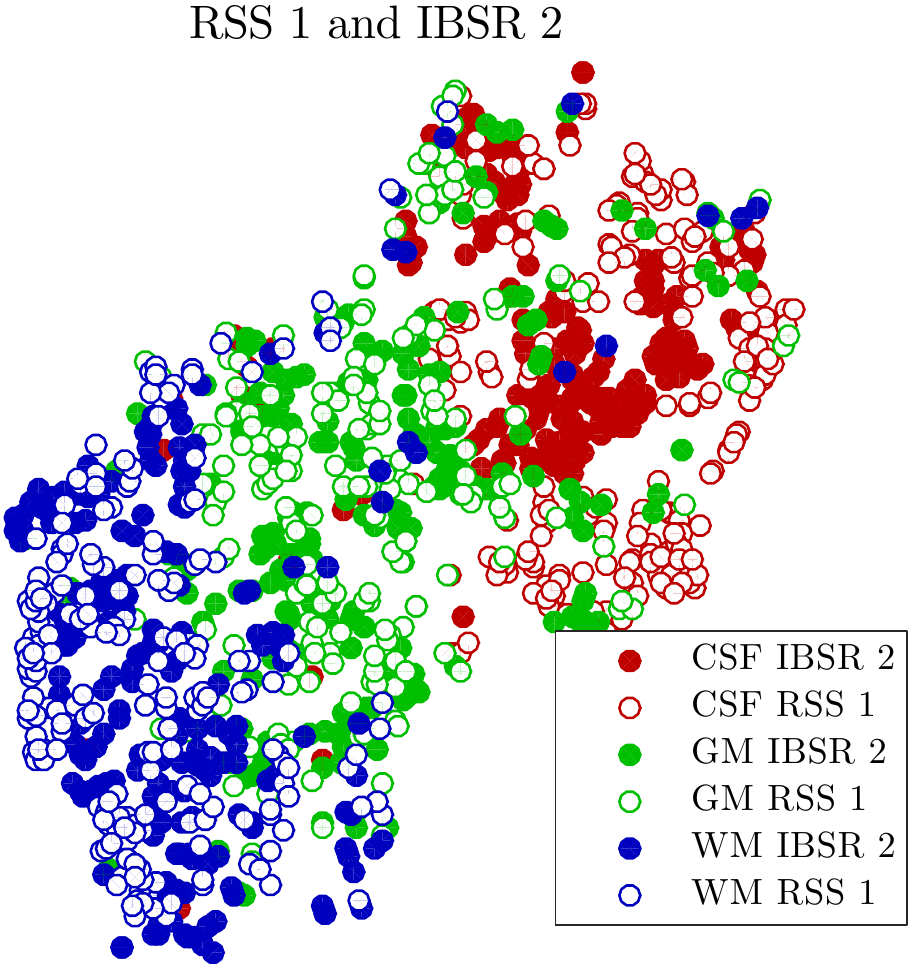}%

\includegraphics[width=0.85\columnwidth]{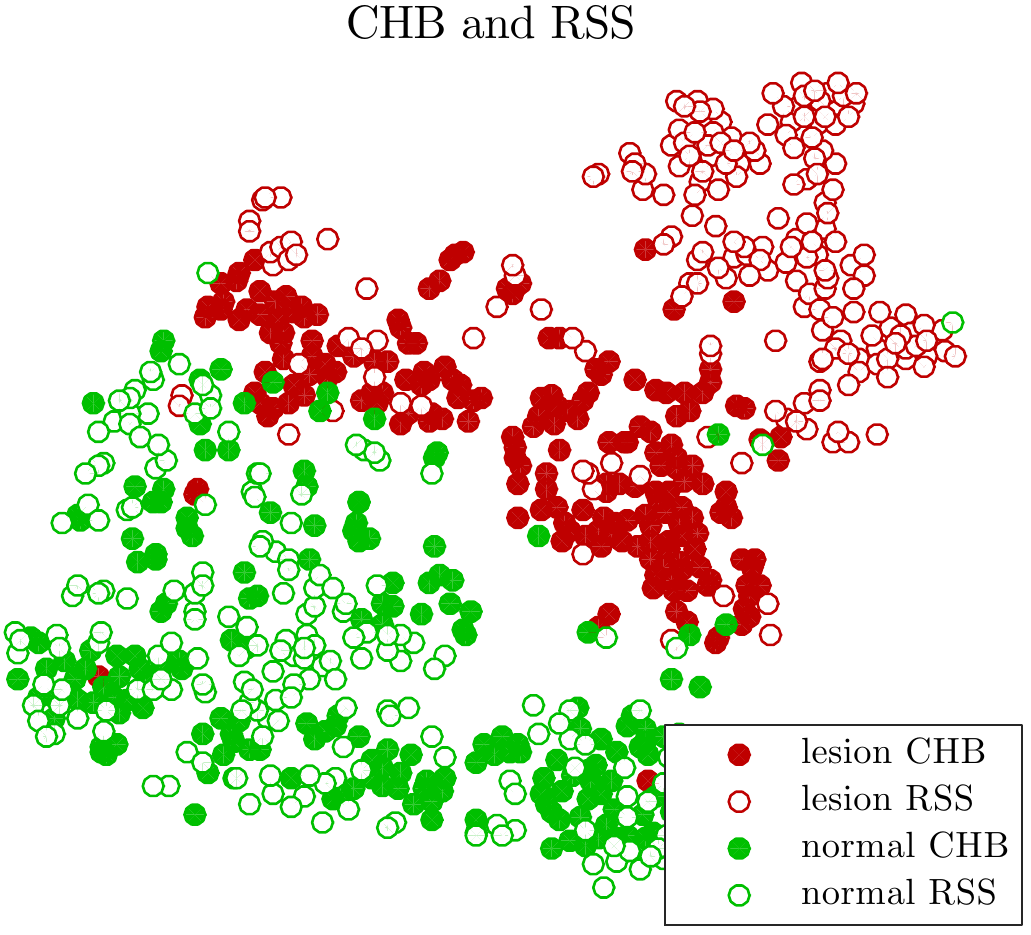}%
\caption{Visualisation of voxels from different-study images in the BT (top) and WML (bottom) segmentation task. After initial normalization, 600 voxels per image are uniformly sampled from 2 images, each from a different source, and their feature vectors are computed. Then a 2D t-SNE~\cite{van2008visualizing} embedding of the feature vectors is performed for visualisation. For a classifier to perform well, voxels of the same class, but from different images, should be close together, but this is not always the case here. For the BT task, note the area in the top right where clusters of CSF voxels from the two images are quite dissimilar. For the WML task, the clusters of lesion voxels from different images almost do not overlap.}%
\label{fig:embedding}%
\end{figure}

For the BT task, we used 13 features: intensity, \{intensity, gradient magnitude, absolute value of Laplacian of intensity\} each after convolution with a Gaussian kernel with $\sigma = 1, 2, 3$ mm$^3$, and the 3D position of the voxel normalized for the size of the brain. To illustrate that despite the initial normalization, these features result in slightly different distributions for different tissue types, we show a 2D embedding of a subset of voxels from two different datasets in Fig.~\ref{fig:embedding} (top).

For the WML task, we used 10 features per channel: intensity, \{intensity, gradient magnitude and Laplacian of Gaussian\} each after convolution with a Gaussian kernel at scales $\{0.5, 1, 2\}$ mm$^3$, resulting in 30 features in total. Each voxel is associated with a binary label, either non-WML or WML. An illustration of how the distributions are different in different sources is shown in Fig.~\ref{fig:embedding} (bottom).

\subsection{Weighted Ensemble Classifier}
 We use the voxels of each training image to train a random forest~\cite{ho1998random,breiman2001random} (RF) classifier, but the method is applicable to other supervised classifiers which can output posterior probabilities. We used RF because of its speed, inherent multi-class ability and success in other medical image analysis tasks, such as brain tumor segmentation~\cite{goetz2016dalsa,zikic2014classifier}, ultrasound tissue characterization~\cite{conjeti2016supervised} and WML segmentation~\cite{geremia2011spatial}.

 RF is itself an ensemble learning method. The idea is to combine several weak, but diverse classifiers -- decision trees -- into a strong learner -- the forest. To train each decision tree, the training voxels are first subsampled. The tree is built by recursively adding nodes. At each node, the features are randomly subsampled, and a feature is chosen that splits the voxels into two groups according to a specified splitting measure. A commonly used measure is the decrease in Gini impurity. The Gini impurity of a set of voxels measures how often a randomly sampled voxel would be misclassified, if it was labeled according to the class priors in that set. In other words, impurity is zero if after splitting each group contains voxels of a single class only. The splitting continues until all leaf nodes are pure, or until a maximum allowed depth is reached. Once training is completed, the features that are chosen for the splits, can be used to calculate the overall importance of each feature in the forest.

 At test time, a voxel is passed down each of the decision trees. Due to subsampling of both data and features during training, the trees are diverse, therefore for each tree, the voxel ends up in a different leaf node. The class labels or class label proportions of these leaf nodes are then combined to output a posterior probability for the test voxel.



\begin{figure*}
    \centering
    \includegraphics[width=0.75\textwidth]{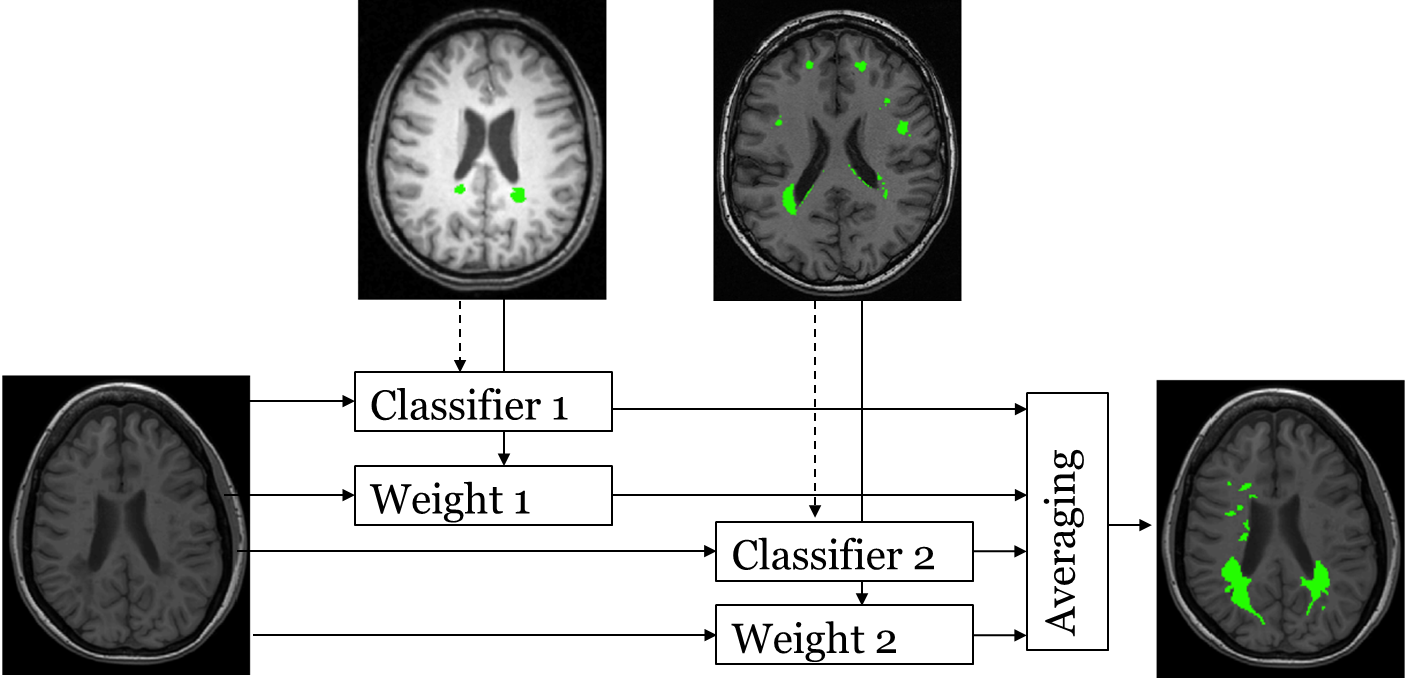}
    \caption{Overview of the method, here illustrated on WML segmentation with 2 training images. At training time (dashed lines) the voxels of each training image are used to train a classifier. At test time (solid lines), the voxels of the test image are classified by each trained classifier, and weights are determined based on the similarity of the test image to the training images. The weighted average of the outputs is the final output of the method.}
    \label{fig:overview}
\end{figure*}

 We classify each voxel by an ensemble of RFs. At test time, our method first computes the distance of the test image to each of the training images as described in Section~\ref{sec:distance}. Each voxel is classified by each of the RF classifiers and the RF outputs are combined with a weighted average rule, where the weights are inversely proportional to the image distances. An overview of the approach is shown in Fig.~\ref{fig:overview}.

Formally, we assume to have access to $M$ training images from various scanners and/or scanning protocols, where the $m$-th image is represented by a set of feature vectors $\{\mathbf{x}^m_i, y^m_i\}$, where $\mathbf{x}^m_i \in \mathbb{R}^n$ is the feature vector describing each voxel and $y^m_i$ is the label indicating the class of the voxel. We do not use information about which scanner and/or scanning protocol each image originates from.

At test time, we want to predict the labels $\{y^{z}_i\}$ of the $z$-th target image with $N_z$ voxels. We assume that at least some of the $M$ training images have similar $p(y|\mathbf{x})$ to to the target image.

The ensemble classifier consists of $M$ base classifiers, where each base classifier $\{f_1, \ldots, f_M\}$ is trained on voxels from a different image, and which can output posterior probabilities. The ensemble decision $F$ is determined by a weighted average of the posteriors $F(\mathbf{x}^z_i) = \frac{1}{M} \sum^M_{m=1} w_{mz} f_m(\mathbf{x}^z_i)$. The weights $w_{mz}$ are inversely proportional to a distance $d_{mz}$ between the images:

\begin{equation}\label{eq:weights}
w_{mz} = (d_{max} - d_{mz})^p / \sum_{m=1}^M (d_{max} - d_{mz})^p
\end{equation}
where $d_{max} = \max_m \{d_{mz}\}$ and $p$ is a parameter that influences the scaling of the weights. With high $p$, similar images get an even higher weight, while dissimilar images are downweighted more. An investigation of this parameter will be presented in Section~\ref{sec:expp}.

In the following section we describe several ways to measure the image distance $d_{mz}$.

\subsection{Image Distances}\label{sec:distance}

In this section we describe measuring the distance $d_{mz}$ between two images, each represented by a set of voxels described in high-dimensional feature space. Ideally, $d_{mz}$ should be small when the images are similar, and thus training a classifier on one image, will lead to good classification performance on the other image. As a sanity check, we therefore also examine a supervised distance measure, which acts as an oracle, as well as three measures which do not use labeled target data. The distance measures are explained below.

\subsubsection{Supervised Distance (Oracle)}

For the oracle distance, we use the target labels to evaluate how well a trained classifier performs on the target image. Instead of using classification error, we use the mean square error (MSE) of the posterior probabilities, because it distinguishes between classifiers that are slightly or very inaccurate.  We denote the posterior probability for class $y$, given by the $m$-th classifier by $f_m^y(\mathbf{x})$. The distance is defined as:

\begin{equation}\label{eq:mse}
d^{sup}_{mz} = \sum_{(\mathbf{x}^z_i, y^z_i)} (1 - f^y_m(\mathbf{x}^z_i))^2.
\end{equation}
We denote this ensemble by $RF^{sup}$.

\subsubsection{Clustering Distance}

In the absence of labels $\{y^z_i\}$, we can estimate the target labels using a clustering procedure. This assumes that per image, the voxels of each class are similar in appearance, i.e. form clusters in the feature space. Here we assume that there are as many clusters as there are classes. By performing clustering and assigning the clusters to the different classes, label estimation is possible. We can thus define $d^{clu}_{mz}$ by performing an unsupervised clustering and replacing the true labels $y^z_i$ by $c^z_i$ in (\ref{eq:mse}), i.e. computing the MSE over the pairs $(\mathbf{x}^z_i, c^z_i)$:

\begin{equation}\label{eq:dclu}
d^{clu}_{mz} = \sum_{(\mathbf{x}^z_i, c^z_i)} (1 - f^c_m(\mathbf{x}^z_i))^2.
\end{equation}

To match the clustering labels to the category labels, prior knowledge about the segmentation task is required. In BT segmentation, this prior knowledge is based on the average (T1) intensity within each cluster. After 3-class unsupervised clustering with $k$-Means, we calculate the average intensity per cluster, and assign the labels \{CSF, GM, WM\} in order of increasing intensity. In WML segmentation, prior knowledge is based on the intensity in the FLAIR scan. After 2-class unsupervised clustering with $k$-Means, we calculate the average intensity per cluster, and assign the labels \{non-WML, WML\} in order of increasing intensity. We use the implementation of $k$-Means from ~\cite{prtools}.

We denote this ensemble by $RF^{clu}$.

\subsubsection{Distribution Distance}

The clustering approach depends both on the classifier and clustering algorithm used. We also propose a classifier-independent approach, where the assumption is that if the probability density functions (PDF) of the source image $P_m(\mathbf{x})$ and target image $P_{z}(\mathbf{x})$ are similar, that the labeling functions $P_m(y|\mathbf{x})$ and $P_{z}(y|\mathbf{x})$ are also similar.  We propose to evaluate the similarity of the PDFs with the Kullback-Leibler divergence, similar to the approach in~\cite{van2015weighting}. A difference is that in~\cite{van2015weighting}, the weights are determined jointly and are used to weight the samples, while we determine the weights individually and use them to weight the classifier outputs.

The divergence distance is defined as:

\begin{equation}\label{eq:div}
d^{div}_{mz} = -\frac{1}{N_z}\sum_{i=1}^{N_z} \log P_m(\mathbf{x}_i^z)
\end{equation}

where $P_m(\mathbf{x})$ is determined by kernel density estimation (KDE) on the samples $\{\mathbf{x}_i^m\}$. We perform KDE with a multivariate Gaussian kernel with width $\Sigma_m^{KL} = \sigma_m^{KL} \cdot I$ where $I$ is the identity matrix. Here $\sigma_m^{KL}$ is determined using Silverman's rule:

\begin{equation}
\sigma_m^{KL} = (\frac{4}{d+2})^{\frac{1}{d+4}} N_m^{\frac{-1}{d+4}}\sigma_m
\end{equation}

where $d$ is the dimensionality of the voxel feature vectors, $N_m$ is the number of voxels and $\sigma_m$ the standard deviation of the voxels. This rule is shown to minimize the mean integrated square error between the actual and the estimated PDF~\cite{silverman1986density}. We denote this ensemble by $RF^{div}$.

\subsubsection{Bag Distance}
Rather than viewing the voxels of each image as a distribution, we can view them as a discrete point set or \emph{bag}. Both the advantage and the disadvantage of this approach is that KDE can be omitted: on the one hand, there is no need to choose a kernel width, on the other hand, outliers which would have been smoothed out by KDE may now greatly influence the results.  A distance that characterizes such bags well even in high-dimensional situations~\cite{cheplygina2015multiple} is defined as:

\begin{equation}\label{eq:bag}
d^{bag}_{mz} = \frac{1}{N_z}\sum_{i=1}^{N_z} \min_j ||\mathbf{x}^z_i - \mathbf{x}^m_j ||^2 .
\end{equation}

In other words, each voxel in the target image is matched with the nearest (in the feature space) source voxel; these nearest neighbor distances are then averaged over all target voxels. We denote this ensemble by $RF^{bag}$.

\subsubsection{Asymmetry of Proposed Distances}

All three of the proposed measures are asymmetric. However, we can only compute both asymmetric versions for $d^{bag}$ and $d^{div}$ because $d^{clu}$ requires labels when computed in the other direction. In (\ref{eq:div}) and (\ref{eq:bag}), we compute the distances from the target samples to the source data ($t2s$). Alternatively, the direction can be reversed by computing distances from the source samples to the target samples ($s2t$). Finally, the distance can be symmetrized, for example by averaging, which we denote as $avg$.

Based on results from pattern recognition classification tasks~\cite{plasencia2013informativeness} and our preliminary results on BT segmentation~\cite{cheplygina2016asymmetric}, our hypothesis is that an ensemble with the $t2s$ similarities outperforms an ensemble with $s2t$ similarities in the opposite direction ($s2t$).

In the $t2s$ distance, all target samples influence the image distance. If some target samples are very mismatched, the image distance will be large. In other words, a high weight assigned to a classifier means that for most samples in the target image, the classifier has seen similar samples (if such samples are present) during training.

\begin{figure*}%
\centering
\includegraphics[width=0.25\textwidth]{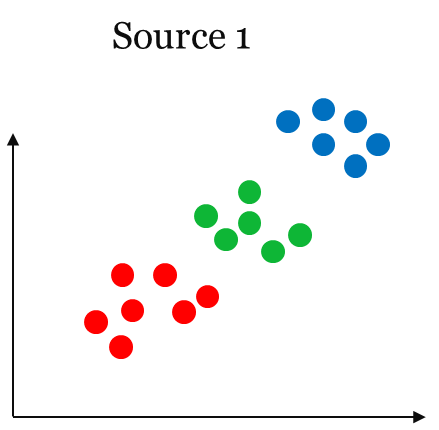}
\includegraphics[width=0.25\textwidth]{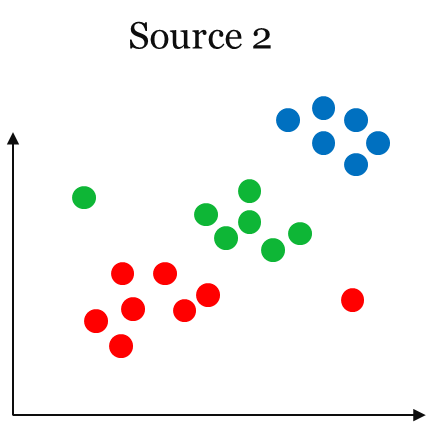}
\includegraphics[width=0.25\textwidth]{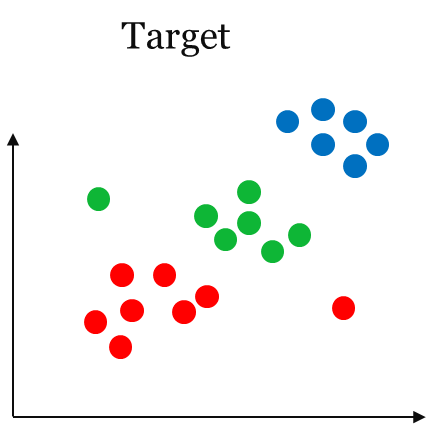}
\caption{Toy example of three images where asymmetric distances can play a role. The average nearest neighbor distance as measured from the source to the target is zero for both sources, while the average nearest neighbor distance as measured from the target to the source is larger for source 1, due to the green and red outliers in the target.}
\label{fig:asymmetry}%
\end{figure*}

On the other hand, if we match source samples to the target samples ($s2t$), these target samples might never be matched, incorrectly keeping the image distance low. Therefore, even if the similarity is high, it is possible that the classifier has no information about large regions of the target feature space. A toy example illustrating this concept is shown in Fig.~\ref{fig:asymmetry}.

The asymmetry of $t2s$ and $s2t$ can be seen as noise that is removed when the distance is symmetrized, for example by averaging ($avg$). If this is the case, we expect $avg$ to outperform $t2s$ and $s2t$. However, if the asymmetry contains information about the task being performed, removing it by symmetrization is likely to deteriorate performance.

\section{Experiments and Results}

In this section we describe the experimental setup for different ways in which we test our method and the corresponding results. First we compare the different image distances in Section \ref{sec:expdist}, followed by a comparison to other competing methods in Section \ref{sec:expcomp}. We then provide more insight into the differences between the image distances and their asymmetric versions. All experiments are conducted on both the BT task with 56 images from four sources, and the WML task with 40 images from three sources.

In all experiments, we use 10,000 voxels per image for training the classifiers, and 50,000 voxels per image for evaluating the classifiers. For BT, we sample these voxels randomly within the brain mask. For WML, we use only a subset of the voxels within the brain mask, following~\cite{van2015weighting}. Because WML appear bright on FLAIR images, we train and test only on voxels within the brain mask with a normalized FLAIR intensity above 0.75. Out of this subset, we sample the voxels in two ways. For training and evaluating the classifiers, we oversample the WML class, such that WML voxels are 10 times more likely to be sampled than non-WML voxels. For calculating the distances at test time when target labels are not available, the voxels are sampled randomly.

The proposed classifier used for both tasks is the same: a random forest (RF) classifier\footnote{https://code.google.com/archive/p/randomforest-matlab/} with 100 trees and otherwise default parameters (sampling with replacement, feature subset size of $\sqrt{n}$ where $n$ is the number of features). Based on our preliminary results on BT segmentation~\cite{cheplygina2016asymmetric}, we use weight scaling parameter $p=10$ for both BT and WML segmentation tasks. This choice ensures that relatively more weight is given to the most similar images; an analysis of this will be provided in Section \ref{sec:expp}.

Following~\cite{opbroek2014transfer,van2015weighting}, we use the percentage of misclassified voxels as the evaluation measure.

\subsection{Comparison of Image Distances} \label{sec:expdist}

We first investigate the effect of the choice image distance $d_{mz}$ on the classifier. Here we compare an ensemble with uniform weights $RF^{uni}$, the three unsupervised distances $RF^{bag}$, $RF^{div}$ and $RF^{clu}$, as well as the oracle $RF^{sup}$, which gives optimistically biased results because the weights are determined using the test image labels. For $RF^{bag}$ and $RF^{div}$, we examine their asymmetric and symmetrized versions.

\begin{figure}
    \centering
    \includegraphics[width=0.45\textwidth]{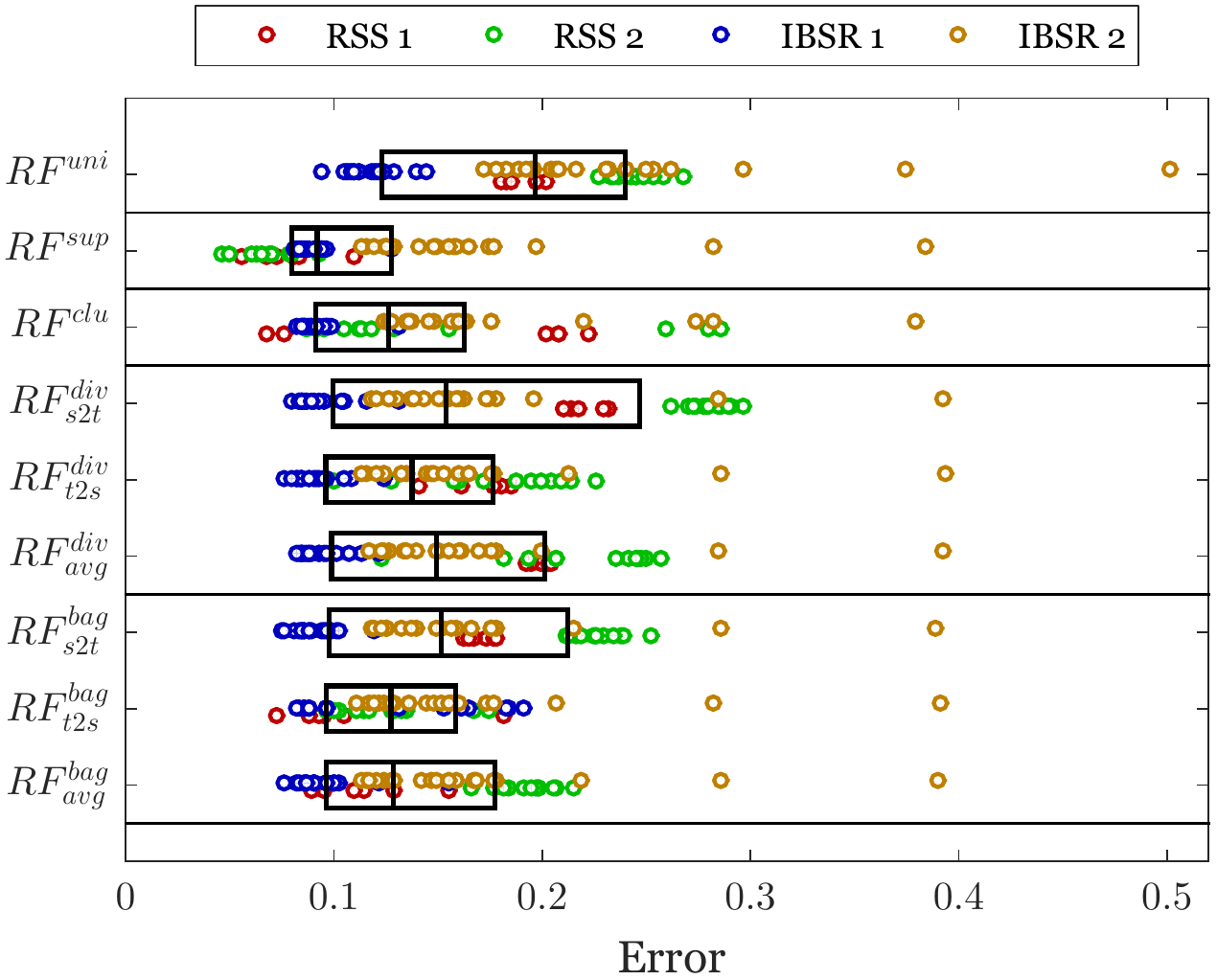}
    \includegraphics[width=0.45\textwidth]{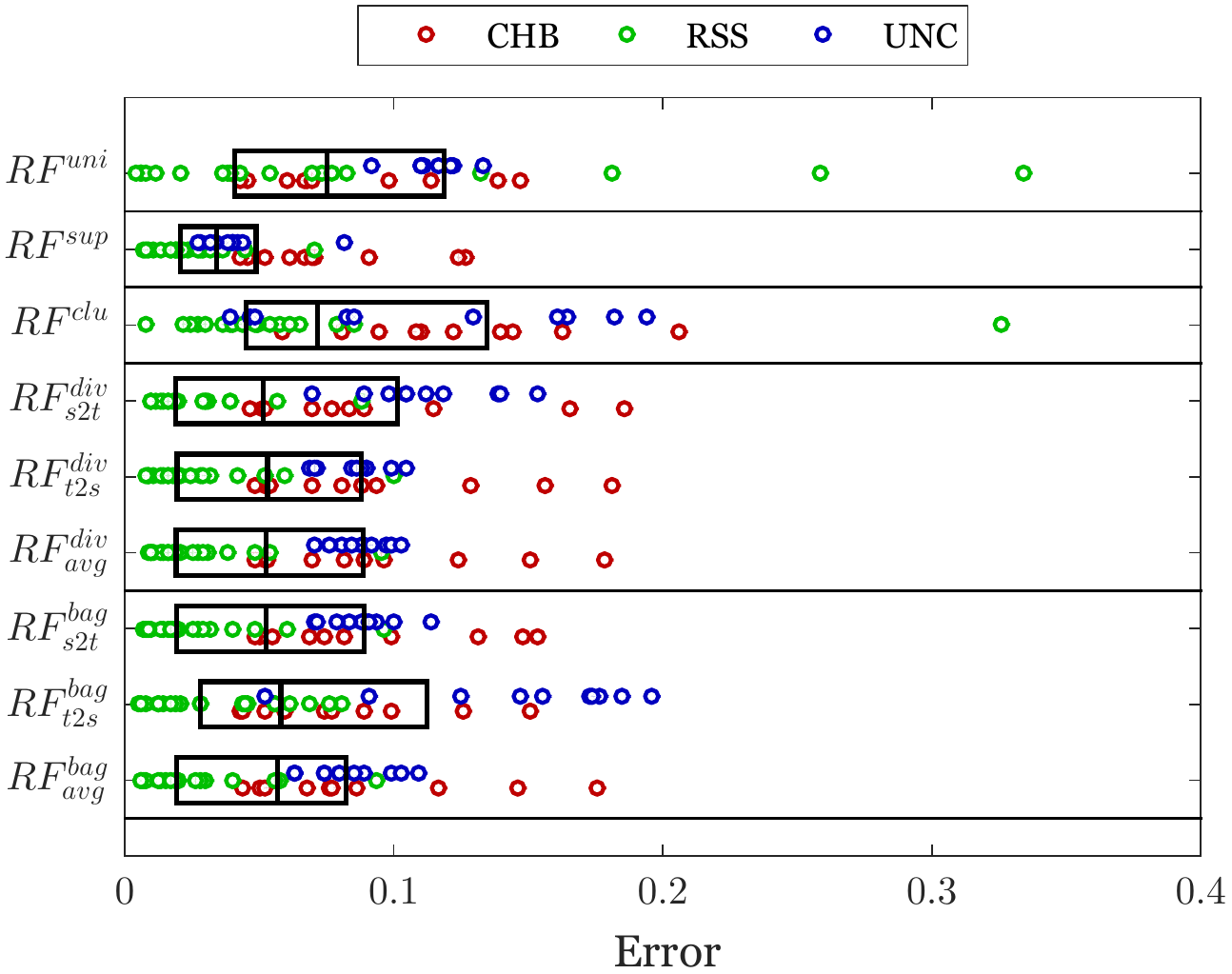}
    \caption{Classification errors for BT (top) and WML (bottom) tasks. Rows correspond to different weighting techniques and baselines: uniform weights $RF^{uni}$, oracle weights $RF^{sup}$, clustering weights $RF^{clu}$, $RF^{div}$ (rows 4-6) and  $RF^{bag}$ (rows 7-9). Each boxplot shows the overall classification errors, while different colors indicate test images from different studies.}
    \label{fig:perfplot}
\end{figure}

The error rates of the different weight strategies are shown in Fig. \ref{fig:perfplot}. The performances of the oracle $RF^{sup}$ demonstrate that with suitable weights, very good performances are attainable. Note that $RF^{sup}$ is an oracle since it uses the target labels, and is only presented in order to get an impression of the best possible performances. For example, these results demonstrate that in the BT experiment, study IBSR 2 has two very atypical images, which cannot be classified well even if supervised weights are used.

Out of the unsupervised similarities, $RF^{clu}$ performs quite well on the BT task, but poorly on the WML task. To understand this result we examine the estimation of the labels by the clustering procedure alone, i.e. matching each cluster with a class label, and assigning that label to all voxels belonging to this cluster. For the BT task, the median error is 0.23, which is worse than most other methods. However, the estimated labels still prove useful in assessing the similarity, because $RF^{clu}$ achieves better results than clustering alone. On the WML task, the clustering procedure alone has a median error of 0.46, which is very poor. Due to the low numbers of lesion voxels, the clustering procedure is not able to capture the lesion class well.  

In the BT task, $RF^{bag}$ gives the best results overall. The asymmetric versions of $RF^{bag}$ and $RF^{div}$ show similar trends. As we hypothesized, measuring the similarity from the target to the source (\emph{t2s}) samples, as in $RF^{bag}_{t2s}$ and $RF^{div}_{t2s}$, outperforms the opposite direction.

In the WML task, the situation with respect to asymmetry is different. All three versions ($t2s$, $s2t$ and $avg$) have quite similar performances, but \emph{t2s} is not the best choice in this case. In particular, with $RF^{bag}_{t2s}$, the results are very poor on UNC. This can be explained by the low prevalence of lesions in this dataset. As only a few voxels in the target images are lesions, the \emph{t2s} image distances are influenced only by a few lesion voxel distances, and therefore are noisy. On the other hand, when \emph{s2t} and therefore \emph{avg} are used, the image distances benefit from relying on a larger set of source lesion voxels.

Based on these results, we choose $RF^{bag}_{t2s}$ for subsequent experiments with the BT task and $RF^{bag}_{avg}$ for the WML task. 

\subsection{Comparison to Other Methods} \label{sec:expcomp}

We compare the weighted ensemble with two baselines and with previous methods from the literature. The baselines are a single RF classifier trained on all source images ($RF^{all}$) and an ensemble with uniform weights for each classifier ($RF^{uni}$). The other competing methods depend on the task and are described below.

For the BT task, we compare our approach to the brain tissue segmentation tool SPM8~\cite{ashburner2005unified} and a weighted SVM~\cite{van2015weighting} (WSVM), which weights the training images by minimizing the KL divergence between training and test data, and trains a weighted SVM. Note that WSVM weights the images jointly, while we weight the classifiers on an individual basis. The results are shown in Table~\ref{tab:exp}.

Comparing to SPM8 and WSVM, our approach is the only one that provides reasonable results for all the four studies. When averaging over all the images, $RF^{bag}_{t2s}$ is significantly better than the other approaches.



\begin{table*}[ht]
\begin{center}

\begin{tabular}{l*{5}{c}}
& \multicolumn{5}{c}{Target study} \\

Method &       RSS1 &      RSS2 &       IBSR1 &  IBSR2 & All \\
\hline
$RF^{all}$ & {\bf 9.5 (2.3)} & 13.1 (1.1) & 22.2 (2.7) & 6.7 (8.4) & 20.5 (8.2) \\
$RF^{uni}$ & 19.1 (1.0) & 24.5 (1.2) & 11.6 (1.3) & 23.7 (7.6) & 19.5 (7.3) \\
$RF^{bag}_{t2s}$ &    {\bf 11.5 (4.2)} &  12.8 (2.6) & {\bf 11.5 (3.9)} & {\bf 16.3 (6.7)} & {\bf 13.5 (5.3)} \\
SPM8  & 12.6 (2.0) & {\bf 10.0 (2.5)} & 20.8 (3.4) & 24.6 (2.1) & 18.9 (6.4) \\
WSVM & 20.3 (4.9) & 16.7 (2.6) & {\bf 10.6 (1.2)} & {\bf 16.2 (6.6)} & 14.9 (5.4) \\
\end{tabular}

\caption{Classification errors (mean and standard deviation, in \%) of different-study methods on BT segmentation. Last column shows average over all 56 images. Bold = best or not significantly worse (paired t-test, $\alpha<0.05$) than best.}
\label{tab:exp}
\end{center}
\end{table*}

For the WML task, we compare our approach to the WSVM. The results are shown in Table~\ref{tab:errtab}. Our approach always outperforms training a single classifier and outperforms uniform weights for RSS and UNC, while having on par performance for CHB. Compared to WSVM, our methods performs on par for CHB, better for RSS and worse for UNC. However, when considering all 40 images, our result significantly outperforms all other methods.

\begin{table*}[ht]
\begin{center}
\begin{tabular}{l*{4}{c}}
& \multicolumn{4}{c}{Target study} \\
    Method &       CHB &       RSS &       UNC & All \\

\hline
$RF^{all}$       & 9.5 (3.4)& 3.4 (1.5)& 18.6 (1.9)& 8.7 (6.7)\\
$RF^{uni}$       & {\bf 8.5 (3.7)}& 7.6 (8.8)& 11.5 (1.1)& 8.8 (6.6)\\
$RF^{bag}_{avg}$ & {\bf 8.9 (4.4)}& {\bf 2.8 (2.3)}& 8.4 (1.6)& {\bf 5.7 (4.1)}\\
WSVM             & {\bf 8.9 (4.6)}& 7.5 (6.7)& {\bf 5.1 (1.1)}& 7.3 (5.4)\\

\end{tabular}
\caption{Classification error (mean and standard deviation, in \%) of different-study methods on WML segmentation. Last column shows average over all 40 images. Bold = best or not significantly worse (paired t-test, $\alpha<0.05$) than best.}
\label{tab:errtab}
\end{center}
\end{table*}

\subsection{Feature Importance} \label{sec:expfeat}

Based on the RF ability to determine feature importance, we examine what features were deemed important when training the source classifiers, and how weighting the classifiers affects the feature importance.

Note that due to the splitting criterion used to determine importance, decrease in Gini impurity, feature importances are generally not independent. For example, in presence of two correlated features $i$ and $j$, if $i$ is always chosen for splits instead of $j$, only the importance of $i$ would be high. However, this is unlikely to occur with a large number of trees, and a relatively small total number of features. We empirically verified whether this could happen in our datasets by comparing the feature importances below with feature importances of a classifier, trained without the most important feature. The correlations were above 0.9, indicating that feature correlations did not have a large influence on determining feature importance.


As the classifiers are trained per image, each classifier has its own feature importances associated with it. We examine average importances for a randomly selected target image. We compare several alternatives of how the importances are averaged: (i) training an ensemble on all other same-study images and averaging the importances, which reflects the best case scenario, (ii) training an ensemble on all different-study images and averaging the importances with uniform weights (same weights as $RF^{uni}$), and training on all different-study images and averaging the importances with the weights given by the proposed method (same weights as  $RF^{bag}$).

\begin{figure*}
    \centering
    \includegraphics[width=0.75\textwidth]{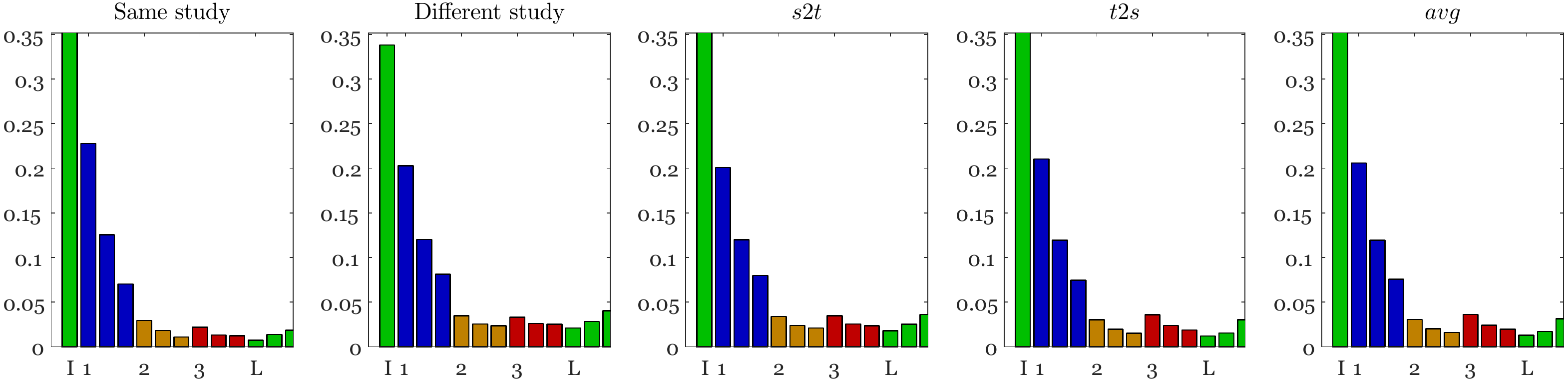}
    \caption{Relative feature importance of the RF ensemble for the BT task,for RSS1. I is the intensity, 1, 2 and 3 represent the features (intensity, gradient magnitude, absolute value of Laplacian) at scales 1mm$^3$, 2mm$^3$ and 3mm$^3$ respectively, and L are the location features. Columns show different strategies:  training  on other same-study images and using uniform weights (best case scenario), training on all different-study images and using uniform weights, or weights from the $s2t$, $t2s$ and $avg$ bag distance.}
    \label{fig:featimpBT}
\end{figure*}

\begin{figure*}
    \centering
    \includegraphics[width=0.75\textwidth]{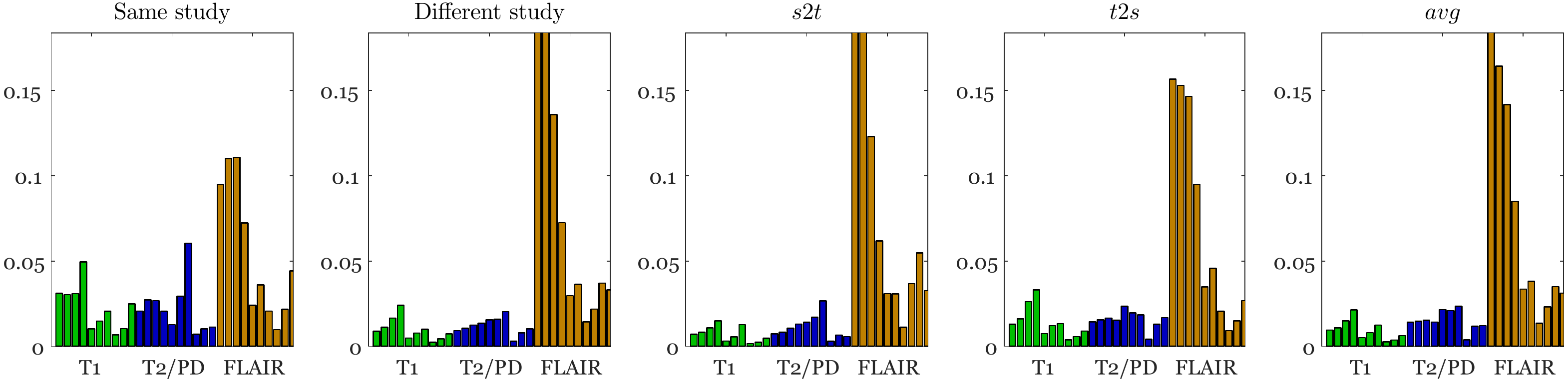}
    \includegraphics[width=0.75\textwidth]{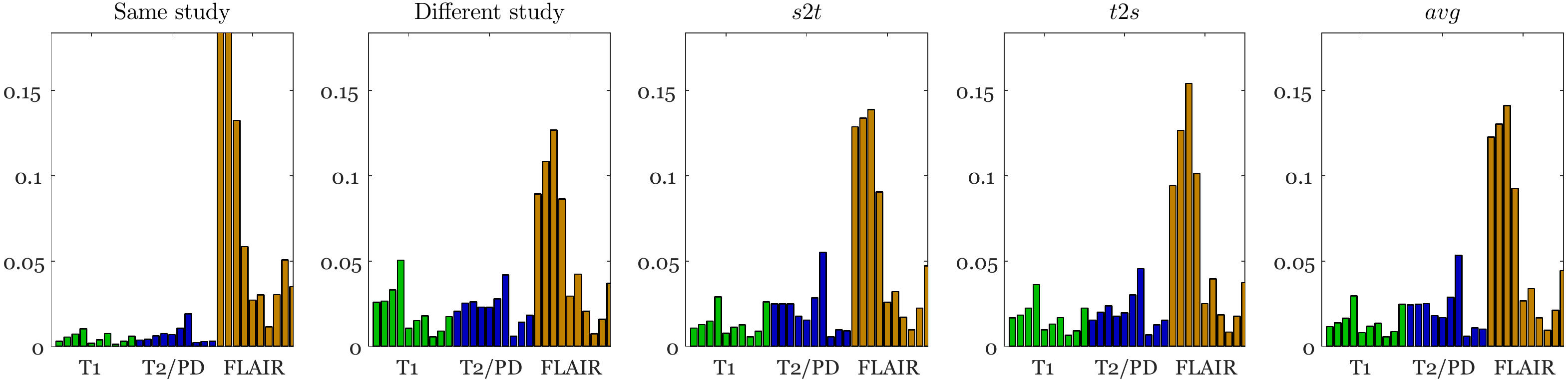}
    \includegraphics[width=0.75\textwidth]{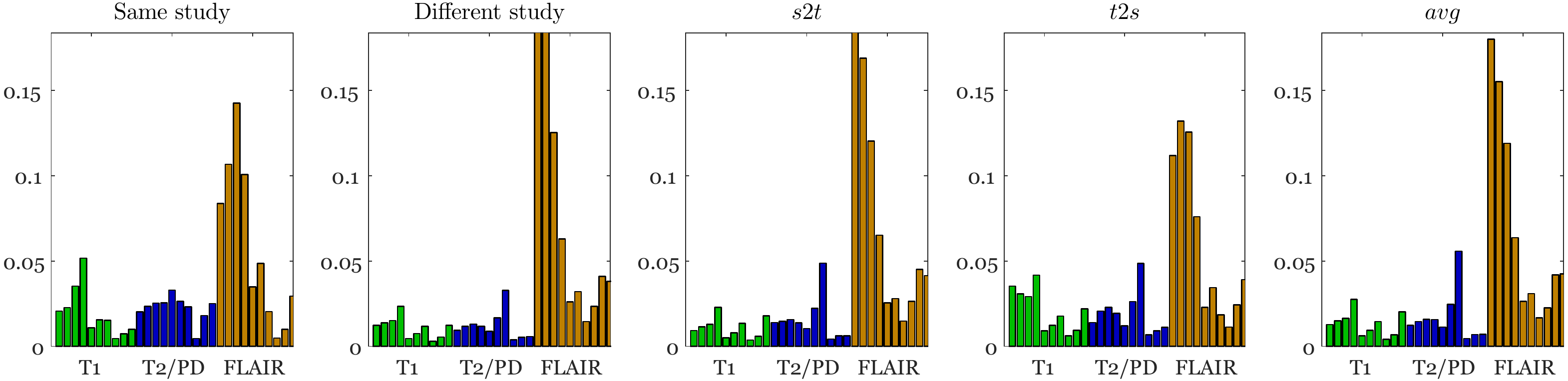}
    \caption{Relative feature importance of the RF ensemble for the WML task for CHB (top), RSS (middle) and UNC (bottom). On the x-axis, T1, T2/PD and FLAIR indicate the features (intensity, gradient magnitude, absolute value of Laplacian) of each modality. Columns show different strategies:  training  on other same-study images and using uniform weights (best case scenario), training on all different-study images and using uniform weights, or weights from the $s2t$, $t2s$ and $avg$ bag distance.}
    \label{fig:featimpWML}
\end{figure*}

For the BT task, the importances are shown in Fig.~\ref{fig:featimpBT}. The relative importance of the features is very similar across datasets, therefore we show the intensities only when RSS1 is the target study. Intensity is the most important feature, followed by features extracted at the smallest scale, and then by the three other sets (features extracted at two larger scales and location features), which are on par with each other. In the ``Different study'' plots, the importance of intensity is slightly lower, but all weighting strategies help to restore this, i.e. columns 3-5 are more similar to the ``Same study" situation.

For the WML task, the importances are shown in Fig.~\ref{fig:featimpWML}. Here the FLAIR features are the most important, followed by T2/PD and T1. The FLAIR features are the most important for RSS, but less so for CHB and UNC. Here the differences between weighting strategies are larger than in the BT task. This can be seen in CHB and UNC, where $t2s$ brings the importances closer to the ``Same-study'' plots, while $s2t$ and $avg$ look very similar to the ``Different study'' plots. This suggests that $t2s$ might be a more logical choice than $s2t$ or $avg$, although in this case this is not reflected in the classifier performances.



\subsection{Weight Scaling}~\label{sec:expp}

\begin{figure}
    \centering
    \includegraphics[width=0.9\columnwidth]{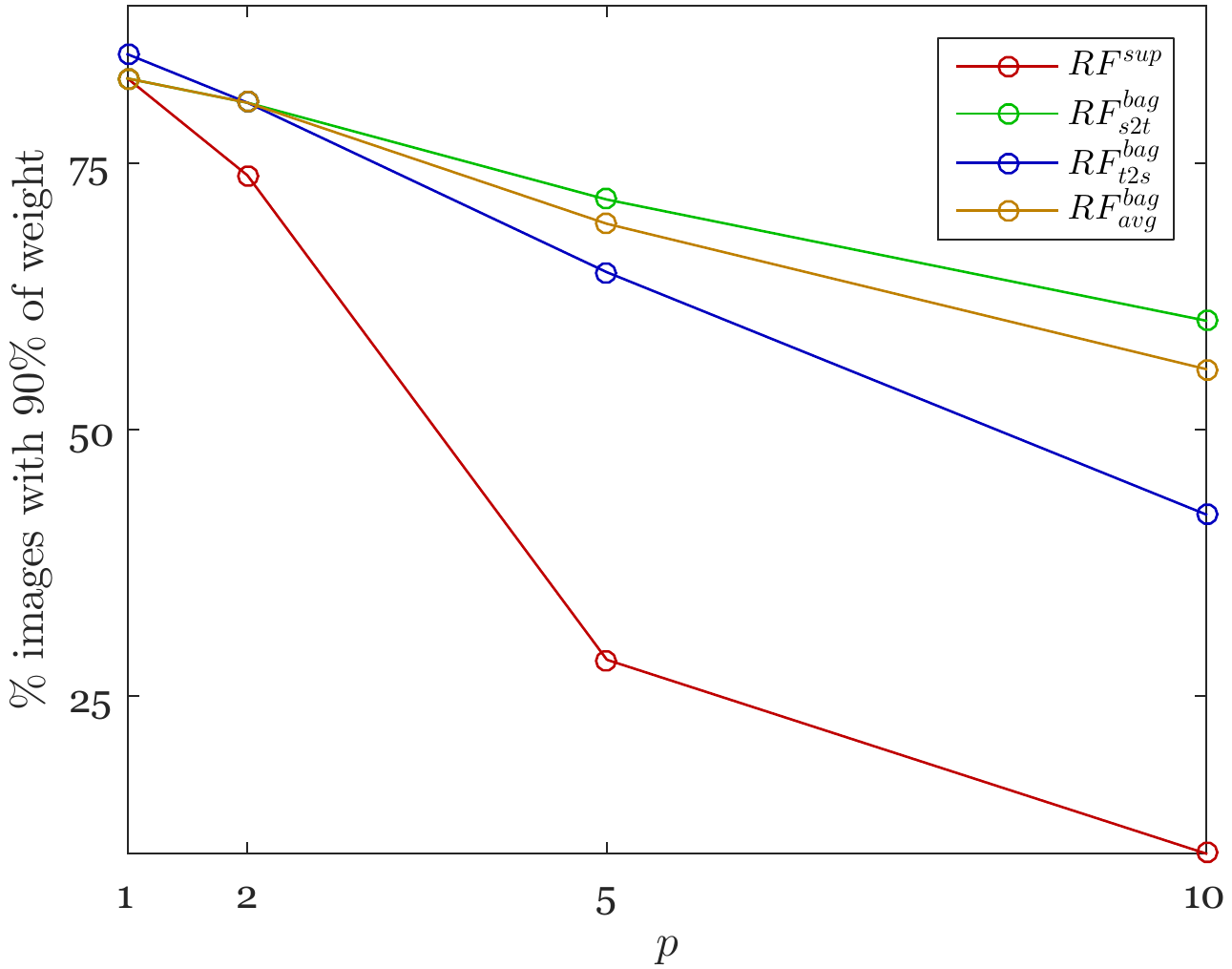}
    \includegraphics[width=0.9\columnwidth]{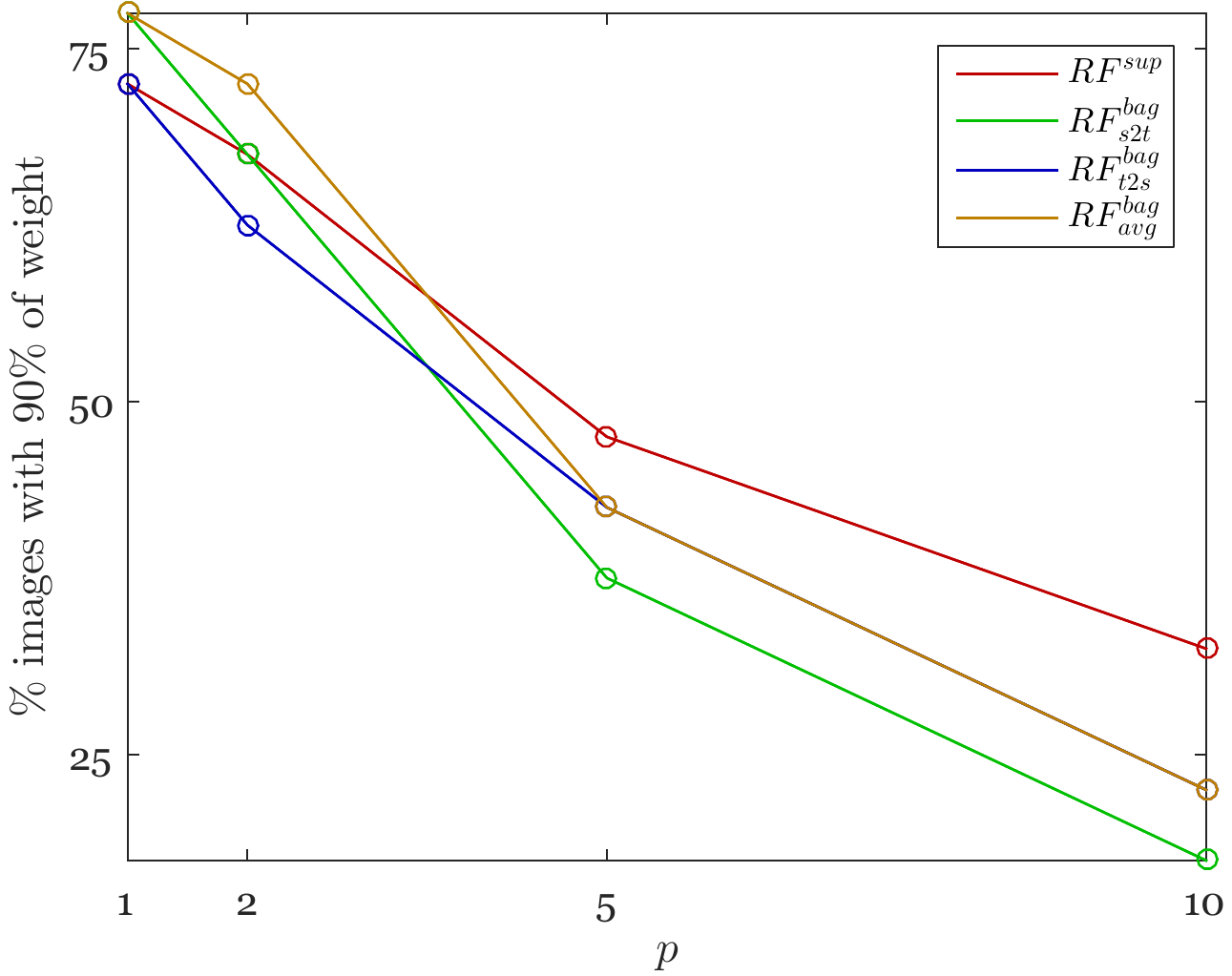}
    \caption{\% of classifiers that receive 90\% of total weight, as a function of scaling parameter $p$ for BT (top) and WML (bottom). Higher \% means the weights are more uniformly distributed amongst classifiers, lower \% means a few relevant classifiers are selected.}
    \label{fig:weightscale}
\end{figure}

Here we examine the effect of the weight scaling parameter $p$ on the weights. Fig.~\ref{fig:weightscale} shows what proportion of classifiers receives 90\% of the total weight with different values of $p$. For $RF^{uni}$, this proportion would be 90\%, as all classifiers have equal weights. With low $p$, the ensembles $RF^{sup}$ and $RF^{bag}$ are very similar to $RF^{uni}$, and most classifiers have an effect on the ensemble. With a larger $p$, the differences in classifier weights become more pronounced, and less classifiers are responsible for the decisions of the ensemble. In other words, a higher $p$ translates into selecting a few most relevant classifiers.

\begin{figure}
    \centering
    \includegraphics[width=0.9\columnwidth]{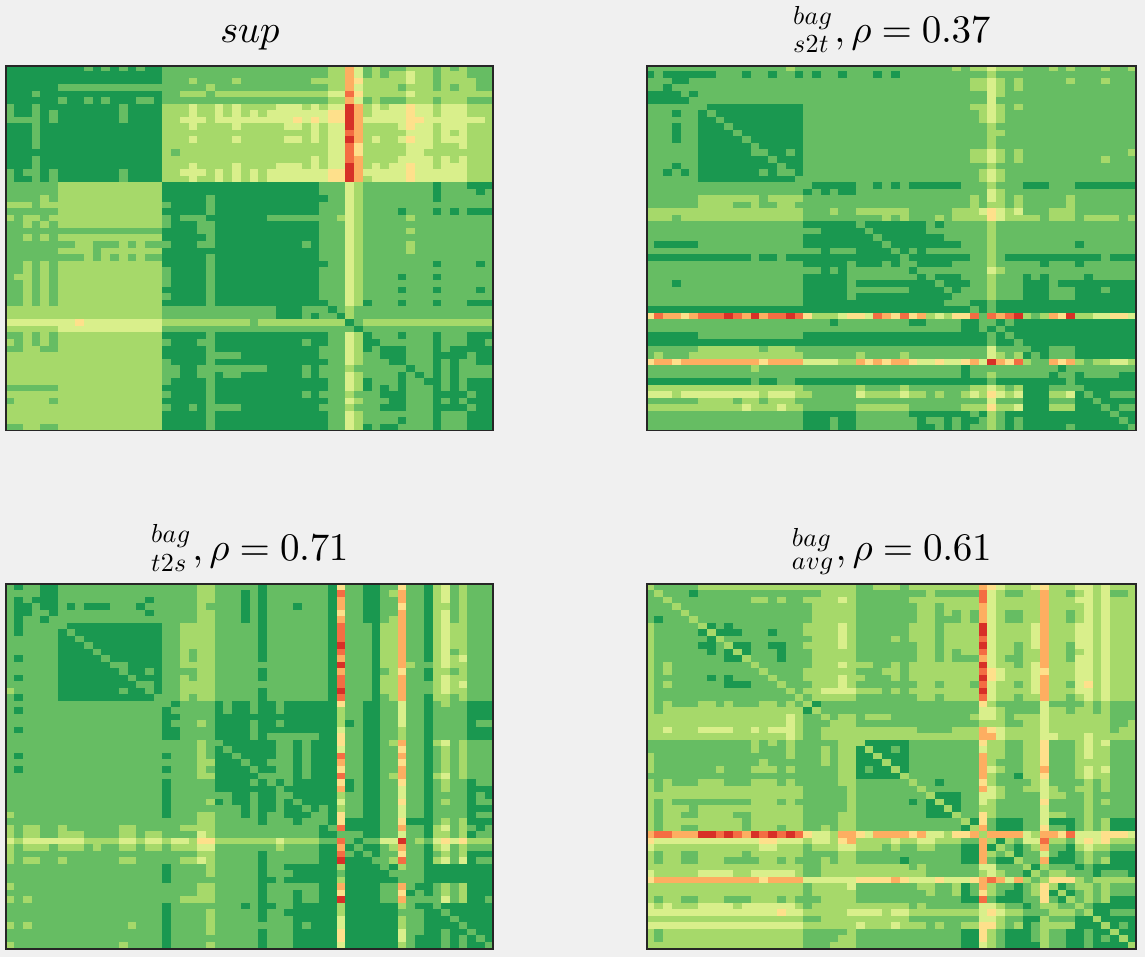}
    \includegraphics[width=0.9\columnwidth]{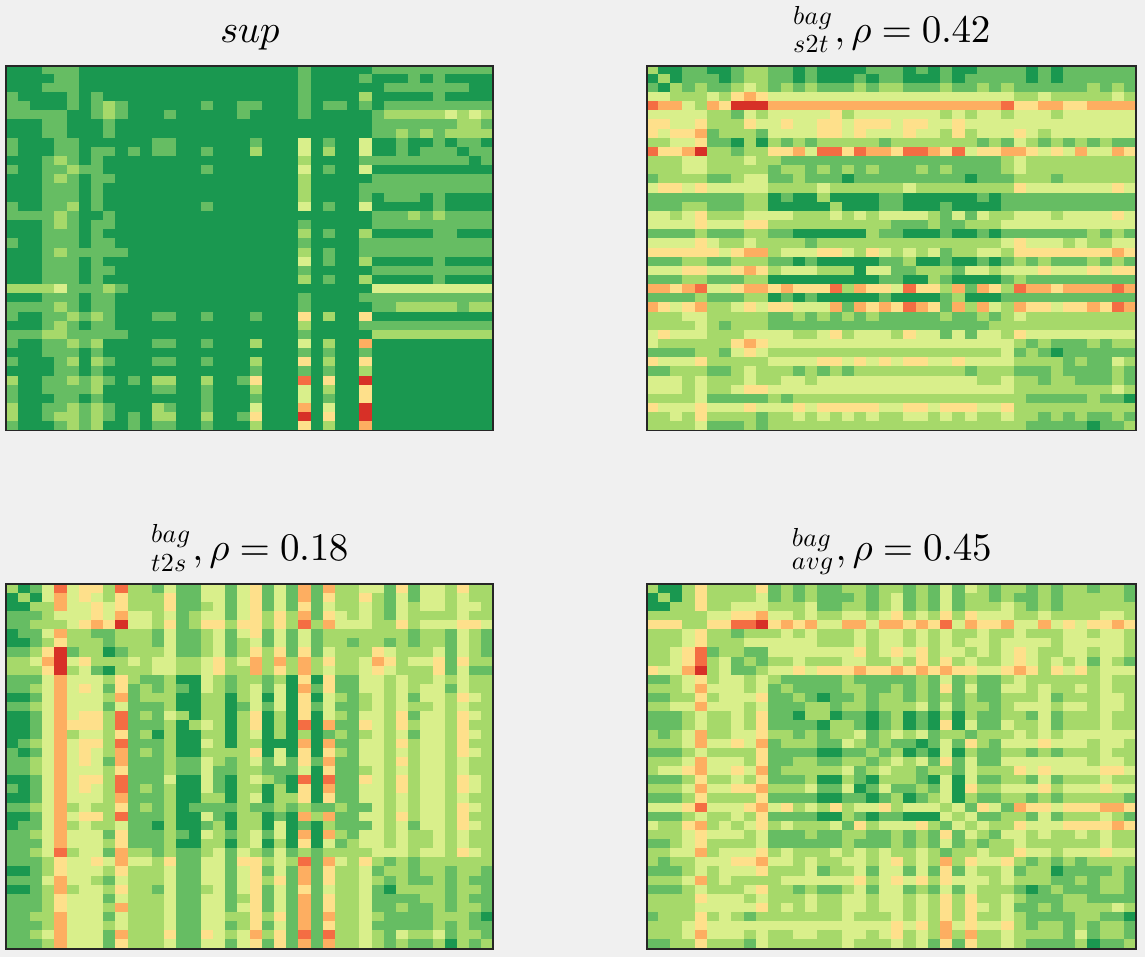}
    \caption{Visualization of oracle $d^{sup}$ and three versions of $d^{bag}$ for BT (top) and WML (bottom). Green = low distance, red = high distance. For $d^{bag}$, the diagonal elements are equal to zero, but for better visualization have been set to the average distance per matrix. $\rho$ shows the average Spearman coefficient between the bag distance and the oracle distance.}
    \label{fig:weightranks}
\end{figure}

Weights influence the performance of the ensemble in two ways: by their ranking and their scaling. Per distance measure, the weights with a different $p$ have the same ranking, but a different scaling, which affects performance. To demonstrate that it is not only a choice of $p$ that leads to our results, in Fig.~\ref{fig:weightranks} we show the distance matrices, from which the weights are computed. For each column, we examine the target image's distances to the source images, and compute the rank correlation between the bag distance and the supervised (oracle) distance. We then average these rank correlations for each distance measure.

A higher coefficient means the method ranks the source images more similarly to the supervised distance, and therefore is likely to perform better. For the BT task, $t2s$ has the highest correlation coefficient, while for WML $avg$ is the best choice. This is consistent with the results we have shown in Section~\ref{sec:expdist}.

\subsection{Computation Time}

To demonstrate the computational efficiency of our method, in this section we present the training and testing times for the proposed approach. The times are indicative, as the code (implemented in MATLAB) was not optimized to reduce computation time. As the classifiers are trained only once, the training time is around 20 seconds per image, which can be done in parallel. Note that the training needs to be done only once, irrespective of the amount of test images. At test time, there are two parts to consider: (i) calculating the distances and (ii) evaluating the trained classifiers on the test image. Calculating the distances is the most time-consuming step. Per test image, the fastest method is $d^{clu}$ (20 seconds), followed by $d^{bag}$ (200 seconds), and by $d^{div}$ (2000 seconds). Evaluation is again fast with around 20 seconds per test image.

\section{Discussion}

We present a weighted RF classifier for BT segmentation and WML segmentation across scanners and scanning protocols. We show robust performances across datasets, while not requiring labeled training data acquired with the target conditions, and not requiring retraining of the classifier. In the following sections, we discuss our results, as well as advantages and limitations of our method in more detail.


\subsection{Differences BT and WML} \label{sec:discussion_datasets}

We tested our methods on datasets from two different tasks, BT and WML. We observed two important differences between the tasks which influenced the performance of the methods, which we discuss in this section. The first difference is the distribution of class priors per task. In BT, the classes are more equally sized than in WML, where the classes are highly imbalanced. The second difference is the heterogeneity of the class (im)balance, or class proportions, in different images. Although in the BT task, the RSS subjects had more CSF than the IBSR subjects, the class proportions across RSS1 and RSS2, or across IBSR1 and IBSR2 was similar. In the WML task, the class proportions different in each subject. Furthermore, source images with similar class proportions were not always available, especially when UNC was the target study.

\begin{figure*}%
\centering
\includegraphics[width=0.45\textwidth]{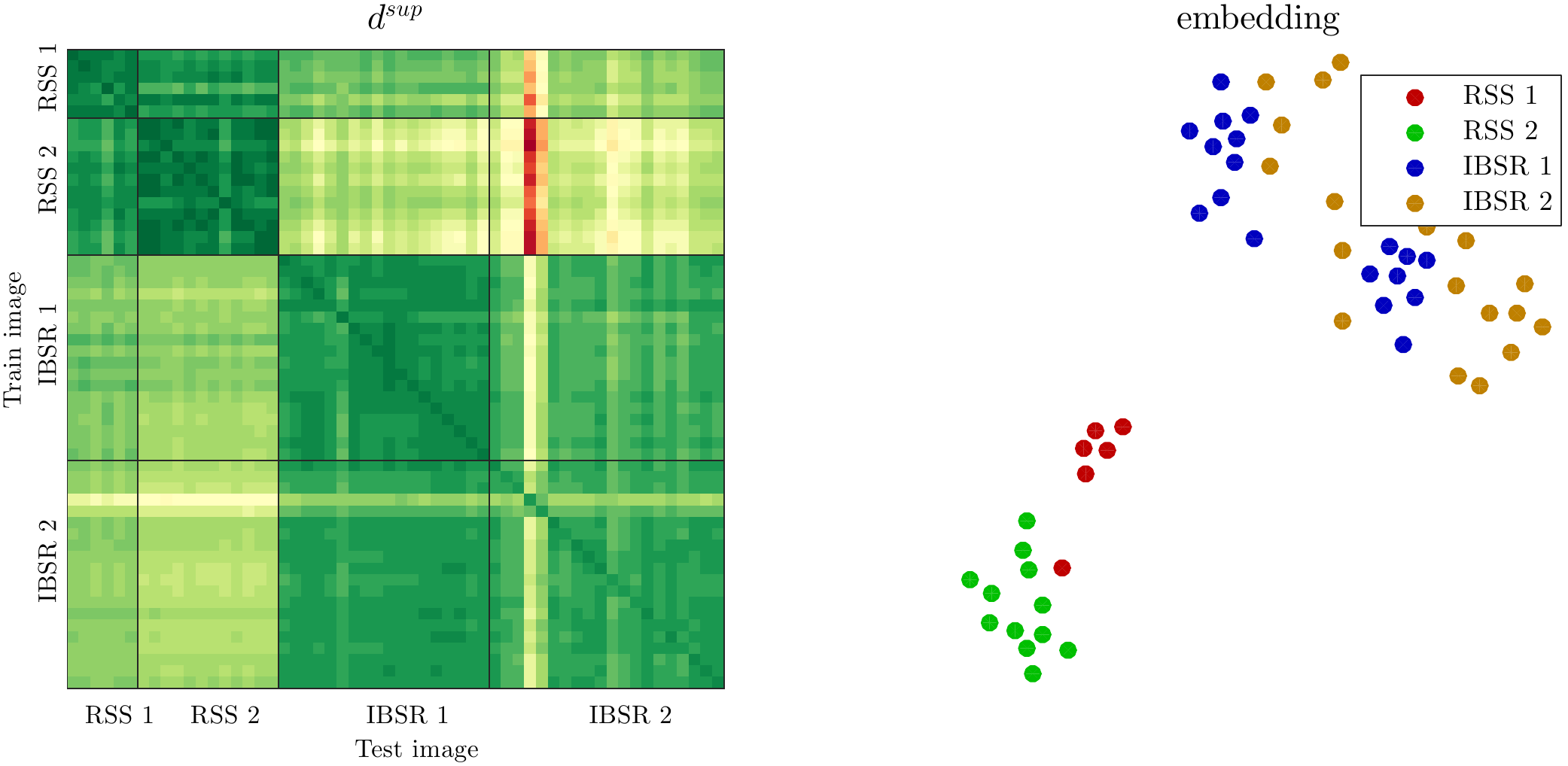}
\includegraphics[width=0.45\textwidth]{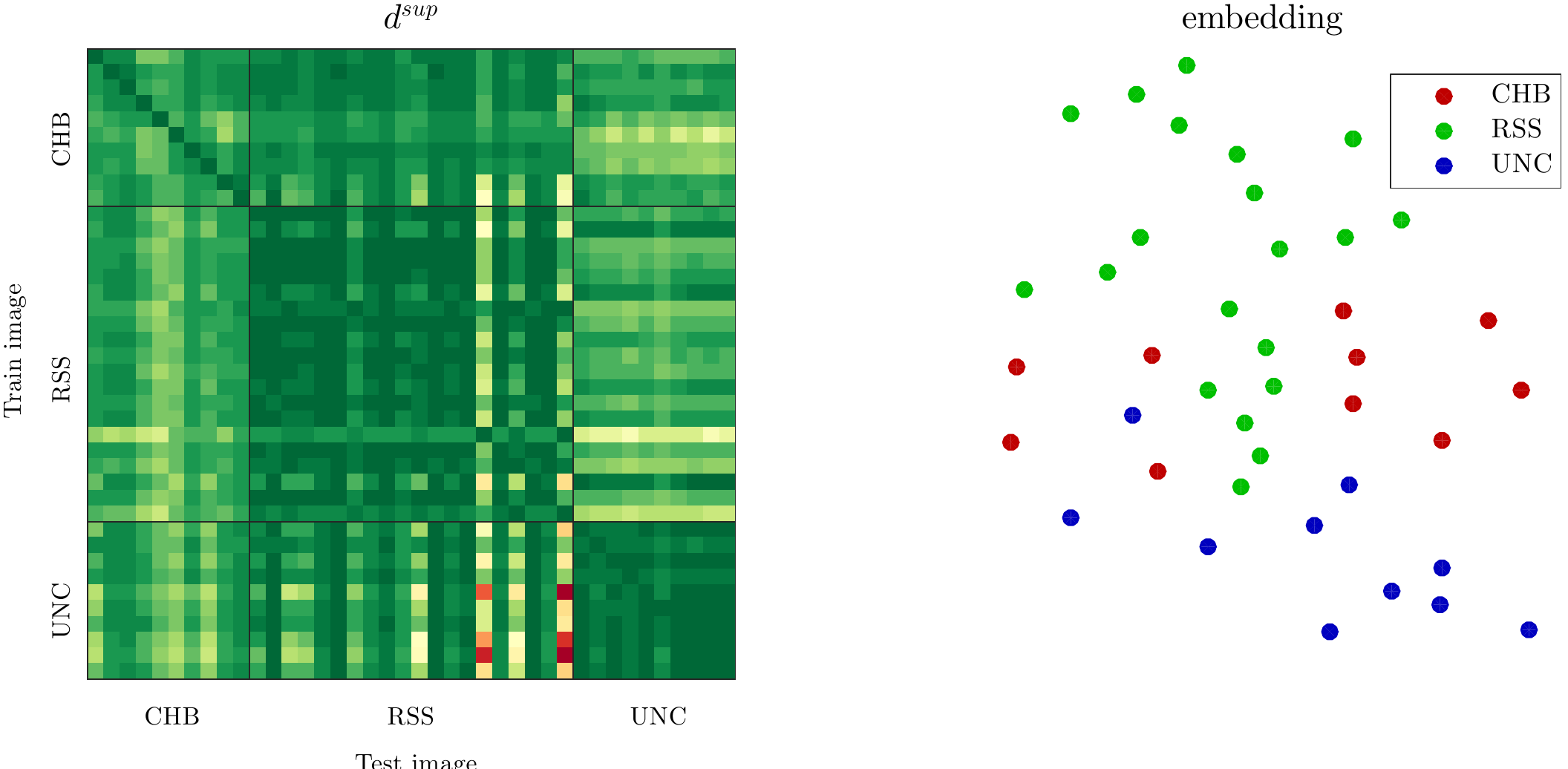}
\caption{Visualizations of the oracle distances $d^{sup}$ (green = low distances/error, red = high distance) and the 2D t-SNE embeddings of these distances for the BT (left) and WML (right) tasks.}
\label{fig:imageembedding}%
\end{figure*}

To better understand the heterogeneity in each task, in Fig.~\ref{fig:imageembedding} we show the supervised distance matrix $d^{sup}$, which shows the performance of each of the classifiers on each of the images, as well as a 2D visualization of the distances in the matrix. In the BT task, both the matrix and the visualization show two clusters: the cluster with RSS1 and RSS2, and the cluster with IBSR1 and IBSR2. This way, for every target image there is always a similar source image available. The situation is different in the WML task. The distances in the matrix are more uniform, and it is less clear what the most similar images are in each case. Although CHB and UNC are using the same scanning protocol, training on an image from CHB and testing on an image from UNC (and vice versa) is not necessarily effective.

In the WML task, UNC is the most dissimilar dataset to the others, demonstrated by the large difference between same-study and different-study performances when UNC is the target study. Because CHB and RSS contain more lesions, our classifier overestimates the number of lesions in UNC, leading to many false positives (FP). This pattern can also be seen in \cite{geremia2011spatial}, where FP rates of several methods are reported. The FP rate can be controlled by adjusting the classifier threshold, and other studies on WML segmentation~\cite{kloppel2011comparison,steenwijk2013accurate} showed that tuning the threshold can improve performance. However,~\cite{kloppel2011comparison} tuned the threshold using training data, which would not help in our case, and \cite{steenwijk2013accurate} tuned the threshold on the test data, optimistically biasing the results.

To investigate whether a different classifier threshold could improve the results in our study, we experimented with an extension of our method, that was informed about the total number of lesion voxels in the target study. We set the threshold such that the total number of voxels classified as lesions is equal to the true total number of lesion voxels in the target study. For CHB and RSS, this threshold was close to the default 0.5 without large changes in performance, but the UNC the informed threshold was much higher, leading to a large improvement in performance. It is a question for further investigation how to set the threshold without using any prior knowledge about the target data.

\subsection{Distance Measures}

For a good classification performance, we need to find source images with $p(y|\mathbf{x})$ similar to that of the target image. In the clustering distance we examined, this is achieved by first estimating the labels $y$ in an unsupervised manner and comparing the $p(y|\mathbf{x})$ of source and target images.  The clustering distance was the most effective for the BT task, but performed poorly on WML because the lesion class could not be captured as a cluster. We expect that using a more sophisticated label estimation procedure would help $RF^{clu}$ achieve better results on the WML task as well. This could be achieved, for example, by initializing the cluster centers at the means of the training data, and constraining the size of the clusters (i.e. that the lesion class is expected to be smaller).

On the other hand, the weights based on the distribution distance and the bag distance assume that $p(y|\mathbf{x})$ is similar when $p(x)$ of the images is similar. The good performances of $RF^{div}$ and $RF^{bag}$ show that this is a reasonable assumption for these datasets. However, it is more appropriate for the BT task, where the classes are more evenly sized than in the WML task where lesion voxels contribute little to $p(\mathbf{x})$.

The distribution distance and the bag distance are two ways to estimate the similarity of $p(\mathbf{x})$, i.e. the distributions of the feature vectors. However, in general similarity can be defined in other ways, for example, by examining the image similarity rather than the feature distribution similarity, or by using properties that are external to the images. For example, in a task of classifying Alzheimer's disease across datasets~\cite{wachinger2016domain}, Wachinger et al used features such as age and sex to weight training images, while the classifier was trained on image features alone. Our weighting strategy takes such characteristics into account implicitly. For example, for the dataset RSS1 with older subjects, older subjects from RSS2 receive higher weights than the younger subjects from IBSR.

It would be interesting to investigate more similarity measures that are unsupervised with respect to the target data. One possibility is STAPLE~\cite{warfield2004simultaneous}, which stands for Simultaneous Truth And Performance Level Estimation. STAPLE takes a collection of candidate segmentations as input and outputs an estimate of the hidden, true segmentation, as well as a performance measure achieved by each candidate, thus giving each candidate a weight. The is the approach taken by~\cite{zikic2014classifier}, who use STAPLE weights for combining classifiers for BT segmentation. However, the output of STAPLE is a consensus segmentation, and would be less appropriate when there are a few similar images, but many highly dissimilar images, as in the WML task.



\subsection{Asymmetry}

An important result is the effect of asymmetry of the similarity measures. On the BT task, measuring the similarity of the target data to the source data ($t2s$) was the best choice, and symmetrizing the similarity deteriorated the results. This supports our hypothesis that $s2t$ ignores important target samples (which are only matched with the $t2s$ distance), and the classifier does not have information about these parts of the target data.

On the other hand, on the WML task $t2s$ was not the best choice in terms of classification error. As we can see in Table~\ref{tab:errtab}, this result was strongly influenced by the results on UNC, where the number of lesions is very low. Because of the low number of lesions, for UNC the $t2s$ distance only includes a few lesion voxels. As such, the lesion voxels do not sufficiently influence the image distances, and $t2s$ was not informative for lesion / non-lesion classification. Matching the larger sets of lesions voxels from the training image to the target data, as in $s2t$ and $avg$, resulted in distances that were more informative.

We used the distances to weight the classifier outputs. Because each classifier has associated feature importances, weighting the classifier outputs also implicitly changes the feature importances of the ensemble. Comparing the weighted feature importances to the best case scenario feature importances (obtained by training on same-study images) also allows us to see which of the weights are more reasonable, i.e. bring the feature importances closer to the best case scenario. In the BT task, the three versions all had a similar effect on the feature importances. However, in the WML task there were noticeable differences, and $t2s$ appeared to be a reasonable measure, even though this was not reflected in the classifier performances. 


\subsection{Limitations}
In this paper we focused on unsupervised transfer learning, assuming that no labeled target data is available. Other recent works on transfer learning in medical image analysis take a different strategy and assume that some labeled target data is available~\cite{conjeti2016supervised,wachinger2016domain}, which may not always be the case. In our method, the absence of labeled target data means that not all differences between the source and target data can be handled. Consider a case where the distributions $p(\mathbf{x})$ of two images are identical, but distributions $p(y|\mathbf{x})$ are very different, for example the decision boundary is shifted and/or rotated. The unsupervised distance measures will output a distance of zero, but the trained classifier will not necessarily be helpful in classifying the target image. Another point where labeled target data would be helpful is setting the classifier threshold, as discussed in \ref{sec:discussion_datasets}.

A limitation of our approach is that it assumes that some sufficiently similar training images are available. This turned out to be a reasonable assumption in our experiments. In the event that none of the training images are similar, the classifier might not be reliable. The classifier could also output the uncertainty along with the predicted label. Such considerations are important when translating classifiers to clinical practice.

A related point is that we consider the similarity of each training image, and thus the accuracy of each classifier independently. However, the performance of the final ensemble depends on two factors: the accuracy of the base classifiers and the diversity of the base classifiers~\cite{kuncheva2003measures}. Therefore, adding only accurate, but not diverse classifiers (i.e. classifiers that all agree with each other) may not be as effective as adding slightly less good classifiers that disagree on several cases. 

\subsection{Implications for other research}
We applied our approach on two segmentation tasks in brain MR images: brain tissue segmentation and white matter lesion segmentation. However, two out of three similarity measures (including the best performing measure) do not use any prior knowledge about brain tissue or about lesions. As such, our approach is not restricted to these applications, and can be applied to other tasks where the training and test distributions are different. We expect our approach to be beneficial when with similar $p(\mathbf{x})$, similar $p(y|\mathbf{x})$ can be expected, and at least some similar training data is available. A example of this situation could be expected in a large, heterogeneous training set.

Likewise, asymmetry in similarity measures is not unique to brain MR segmentation. In previous work, we found asymmetry to be informative when classifying sets of feature vectors in several pattern recognition applications outside of the medical imaging field~\cite{plasencia2013informativeness,cheplygina2015multiple}. The default strategy here would have been to symmetrize the similarities. However, we found that in the BT task, $t2s$ was most effective, and that symmetrizing could deteriorate the results. This suggests that this might be a more widespread issue. Similarities are abundant in medical imaging and are important when weighting training samples, weighting candidate segmentations or classifiers (such as this paper), or even when using a $k$-nearest neighbor classifier. We therefore urge researchers to consider whether asymmetry might be informative in their applications as well.

\section{Conclusions}

We proposed an ensemble approach for transfer learning, where training and test data originate from different distributions. The ensemble is a weighted combination of classifiers, where each classifier is trained on a source image that may be dissimilar to the test or target image. We investigated three weighting methods, which depend on distance measures between the source image and the target image: a clustering distance, a divergence measure, and a bag distance measure. These distance measures are unsupervised with respect to the target image i.e., no labeled data from the target image is required. We showed that weighing the classifiers this way outperforms training a classifier on all the data, or assigning uniform weights to the source classifiers. The best performing distance measure was an asymmetric bag distance measure based on averaging the nearest neighbor distances between the feature vectors describing the voxels of the source and target images. We showed that asymmetry is an important factor that must be carefully considered, rather than noise that must be removed by symmetrizing the distance. We applied our method on two different applications: brain tissue segmentation and white matter lesion segmentation, and achieved excellent results on seven datasets, acquired at different centers and with different scanners and scanning protocols. An additional advantage of our method is that the classifiers do not need retraining when novel target data becomes available. We therefore believe our approach will be useful for longitudinal or multi-center studies in which multiple protocols are used, as well as in clinical practice.

\section*{Acknowledgements}

This research was performed as part of the research project ``Transfer learning in biomedical image analysis'' which is financed by the Netherlands Organization for Scientific Research (NWO) grant no. 639.022.010. We thank Martin Styner for his permission to use the MS Lesion challenge data.

\bibliographystyle{elsarticle-num}

\bibliography{refs}

\end{document}